%% file: sample_paper.tex
\pdfoutput=1
\documentclass[twoside]{article}

\usepackage[preprint]{aistats2027}
\usepackage{amsmath,amssymb,amsfonts}
\usepackage{algorithmic}
\usepackage{graphicx}
\usepackage{textcomp}
\usepackage{xcolor}
\usepackage[breaklinks=true]{hyperref}
\usepackage{url}
\usepackage{booktabs}
\usepackage{multirow}
\usepackage{nicefrac}
\usepackage{microtype}
\usepackage{enumitem}

\usepackage{amsthm}
\usepackage{algorithm}
\usepackage{thmtools}
\usepackage{thm-restate}
\usepackage{tikz}
\usetikzlibrary{arrows.meta, positioning}

\theoremstyle{definition}
\newtheorem{lemma}{Lemma}

\theoremstyle{definition}
\newtheorem{definition}{Definition}

\theoremstyle{plain}

\theoremstyle{plain}
\newtheorem{proposition}{Proposition}

\theoremstyle{definition}

\theoremstyle{definition}

\theoremstyle{remark}
\newtheorem{remark}{Remark}

\theoremstyle{plain}
\newtheorem{corollary}{Corollary}

\usepackage[square,numbers,sort&compress]{natbib}

\begin{document}

\pagestyle{fancy}
\cfoot{\thepage}
\setlength{\footskip}{24pt}
\fancypagestyle{titlepage}{\fancyhf{}\cfoot{\thepage}\renewcommand{\headrulewidth}{0pt}}

\runningtitle{Cooperative Multi-Agent Reinforcement Learning with Submodular Trajectory Rewards}

\twocolumn[

\aistatstitle{Cooperative Multi-Agent Reinforcement Learning\\ with Submodular Trajectory Rewards}

\aistatsauthor{Yasaman Zolfimoselo \And Nikolas Geroliminis \And Maryam Kamgarpour}

\aistatsaddress{EPFL, Switzerland \And EPFL, Switzerland \And EPFL, Switzerland}]

\begingroup
\makeatletter
\renewcommand{\@makefntext}[1]{\noindent\raggedright#1}
\makeatother
\renewcommand{\thefootnote}{}
\footnotetext{Emails: \texttt{\char`\{yasaman.zolfimoselo, \:nikolas.geroliminis, \:maryam.kamgarpour\char`\}@epfl.ch}}
\endgroup

\thispagestyle{titlepage}
\begin{abstract}
We study cooperative multi-agent reinforcement learning with decoupled dynamics and a monotone submodular reward function over joint trajectories. Due to the non-additive reward, an optimal policy may depend on the history of all agents. This dependence makes exact optimization exponential in the horizon and the number of agents. To overcome this challenge while retaining theoretical guarantees, we introduce pre-window curvature $c_m$, which quantifies how states visited more than $m$ steps earlier reduce current marginal gains. Using this curvature to control the error from restricting the policy of each agent to depend only on its last $m+1$ states, we propose a greedy algorithm that exploits the decoupled dynamics to optimize local policies sequentially. For fixed $m$, the algorithm achieves a $\frac{1-c_m}{2-c_m}$ approximation guarantee up to estimation error, and its computation time is polynomial in the number of agents and the horizon. We recover the known $1/2$-factor approximation for time-additive rewards and obtain an approximation factor no smaller than the known total-curvature bound for a single agent. We formulate multi-agent traffic monitoring as an example of this framework and evaluate our algorithm on it.
\end{abstract}

\section{Introduction}
In standard reinforcement learning (RL), a single agent maximizes an expected sum of stage rewards, which depend on the agent's current state and action. Extending the single-agent formulation, cooperative multi-agent reinforcement learning (MARL) studies a team of agents that coordinate their actions to maximize a common objective. Each stage reward then depends on the agents' current joint state and action. A further simplifying assumption is that each stage reward is a sum of the agents' local rewards~\citep{qu2020}. These two forms of additivity separate rewards over time and across agents, respectively. However, in many cooperative applications, the objective is not additive over time or across agents. Without these forms of additivity, an optimal policy may depend on the history of all agents.

In this paper, we consider a common monotone submodular reward defined over the visits along the trajectories of all agents. A visit is a time-state pair $(t,s)$ indicating that at least one agent is in state $s$ at time $t$. Submodularity captures diminishing returns both across agents and over time. This setting is relevant to informative sensing and path planning~\citep{krause2008,singh2009adaptive}, cooperative coverage and biodiversity monitoring~\citep{prajapat2022coverage}, and multi-agent persistent monitoring~\citep{chevaleyre2004,cassandras2012,rezazadeh2021}. Traffic monitoring with multiple unmanned aerial vehicles (UAVs)~\citep{kanistras2013,outay2020,maljkovic2025} is a form of persistent monitoring that we use as a running example. In this example, monitoring a road adds less value if it was monitored recently or if another UAV monitors it at the same time. To study such rewards, we formulate a submodular multi-agent Markov decision process (MDP) with decoupled dynamics, in which the next state of each agent depends only on its own current state and action.

\subsection{Related Work}
\paragraph{Submodular maximization and path planning.}
The greedy algorithm achieves a $(1-1/e)$-approximation for maximizing a monotone submodular set function under a cardinality constraint~\citep{nemhauser1978}. This guarantee improves when the function has total curvature $c<1$~\citep{conforti1984}. These results were extended to settings where the objective function depends on the order of selection~\citep{tschiatschek2017,mitrovic2019,zhang2016string}. In all these works, removing elements from a feasible selection leaves it feasible. In our setting, visits must form trajectories consistent with the dynamics, so removing a visit can make a feasible selection infeasible. Informative path planning maximizes submodular objectives under such constraints~\citep{chekuri2005,singh2009}. For multiple robots, sequential allocation yields a team guarantee when each robot's marginal problem under fixed predecessor policies is solved optimally or with an approximation guarantee~\citep{singh2009,singh2009adaptive,atanasov2015}. These planning models assume deterministic transition dynamics. Under stochastic dynamics, agents must optimize policies that account for uncertain next states.

\paragraph{Submodular RL.}
Submodular RL extends these planning models to stochastic dynamics. In this setting, an agent maximizes an expected monotone submodular reward over visits along its trajectory~\citep{wang2020planning,prajapat2024}. Global RL generalizes submodular RL to broader classes of trajectory rewards and proposes semi-gradient methods for their optimization~\citep{de2024global}. For submodular rewards, constant-factor approximation is generally intractable~\citep{prajapat2024}, but a $(1-c)$-factor approximation is possible with total curvature $c<1$~\citep{prajapat2024,de2024global}. These results are limited to a single agent. Treating a team as one agent makes the joint state space exponential in the number of agents. Moreover, diminishing returns across agents at the same time step can make the $(1-c)$ guarantee vacuous even for time-additive rewards, as we show in Section~\ref{subsec:intractability}.

\paragraph{Submodular MARL.}
In the multi-agent setting, \citet{chen2026} consider time-additive rewards that are monotone submodular across agents at each stage. Under known decoupled dynamics, they optimize the policies sequentially and obtain a $\frac12$-approximation in time polynomial in the number of agents. \citet{liu2026submodular} study stage rewards defined as marginal gains of a submodular trajectory reward, but their guarantee requires every choice to remain feasible regardless of earlier actions. Hence, the guarantees of~\citet{chen2026} and~\citet{liu2026submodular} do not cover rewards with diminishing returns over time when the dynamics constrain future visits.

\paragraph{Multi-agent persistent monitoring.}
Persistent monitoring coordinates mobile agents to revisit important locations~\citep{chevaleyre2004,cassandras2012}. Closest to our setting, \citet{rezazadeh2021} consider rewards that grow with the time since the last visit and plan over a receding horizon under deterministic dynamics. Their $\frac12$-factor approximation is relative to the optimum within each planning horizon, not to the full-horizon optimal return.

To our knowledge, no prior work gives a polynomial-time algorithm with a curvature-dependent guarantee for MARL with submodular trajectory rewards when stochastic dynamics constrain future visits. Our formulation addresses this setting under decoupled dynamics and contains the settings of~\citet{chen2026} and~\citet{prajapat2024} as special cases, as Remark~\ref{rem:special_cases} shows.

\subsection{Contributions}
Building on the above past work, our contributions are summarized as follows.
\begin{enumerate}[itemsep=4pt, topsep=4pt, parsep=0pt]
  \item We introduce a new notion of curvature called pre-window curvature $c_m$, which measures the maximum relative reduction in marginal gains caused by visits before an $m$-step window.
  \item Using this notion of curvature, we modify the objective so that stage rewards depend only on the visits of all agents in the last $m+1$ steps. Jointly maximizing this objective yields a policy that achieves a $(1-c_m)$-approximation of the optimal return.
  \item We develop Greedy Windowed Policy Optimization (GWPO), which sequentially optimizes each agent's policy using marginal rewards based on complete predecessor trajectories and its own $m$-step history. For fixed $m$, it runs in polynomial time and achieves the $\frac{1-c_m}{2-c_m}$ approximation guarantee of Theorem~\ref{thm:sequential_guarantee} relative to the optimal global history-dependent return.
  \item We formulate multi-UAV traffic monitoring as a submodular multi-agent MDP, capturing diminishing returns from repeated monitoring across UAVs and time, and we evaluate GWPO in this setting.
\end{enumerate}

\section{Problem Setting}\label{sec:problem}
In this section, we first introduce the notation. We then recall submodularity and curvature, which our analysis builds on. Finally, we define the submodular multi-agent MDP and the optimization problem, and illustrate the framework through traffic monitoring.
\subsection{Notation}
Throughout the paper, we write $[n] = \{1, \dots, n\}$ for $n \geq 1$, and $[0] = \emptyset$. Bold symbols denote joint quantities over the $N$ agents. For example, $\boldsymbol x$ denotes $(x_1,\ldots,x_N)$. For a function $g$ on a product set $\mathcal X\times\mathcal Y$ and $z=(x,y)$, we write $g(z)$ and $g(x,y)$ interchangeably. Let $\mathcal{D}(\mathcal{X})$ denote the probability simplex over a finite set $\mathcal{X}$. Fix a set $\mathcal S$. For a set $B\subseteq\{0,1,2,\ldots\}\times\mathcal S$ and integers $t$ and $t'$, let
\begin{equation}\label{eq:restriction}
  B_{t:t'}=\{(u,s)\in B : t\leq u\leq t'\}.
\end{equation}
For $t'=t$, we write $B_t$ for $B_{t:t}$. For a set $B_i$ of agent $i$, we write $B_{i,t:t'}$, and for a tuple $\boldsymbol B=(B_1,\ldots,B_N)$ of such sets, we apply this definition componentwise, $\boldsymbol B_{t:t'}=(B_{1,t:t'},\ldots,B_{N,t:t'})$.

\subsection{Submodularity and Curvature}
We recall monotone submodular functions and their total curvature.

\begin{definition}[Monotone submodular function]
\label{def:monotone_submodular}
Let $\mathcal{V}$ be a finite ground set and let $F: 2^{\mathcal{V}} \rightarrow \mathbb{R}_{\geq 0}$ satisfy $F(\emptyset) = 0$. For $X, B \subseteq \mathcal{V}$, the marginal gain of $X$ given $B$ is
\begin{equation*}
\Delta(X \mid B) = F(X \cup B) - F(B),
\end{equation*}
abbreviated as $\Delta(x \mid B)$ when $X = \{x\}$. The function $F$ is \emph{monotone} if $\Delta(x \mid B) \geq 0$ for every $B \subseteq \mathcal{V}$ and $x \in \mathcal{V}$. It is \emph{submodular} if
\begin{equation*}
\Delta(x \mid B) \;\geq\; \Delta(x \mid B')
\quad\text{whenever } B \subseteq B' \text{ and } x \in \mathcal{V}.
\end{equation*}
\end{definition}

Submodularity states that marginal gains decrease as $B$ grows, but it does not quantify how much $\Delta(x \mid B)$ can decrease relative to $\Delta(x \mid \emptyset) = F(\{x\})$. Total curvature measures this worst-case relative decrease.

\begin{definition}[Total curvature]
\label{def:tot_curvature}
The total curvature of a normalized monotone submodular function $F$ is
\begin{equation*}
c = 1-\min_{\substack{B\subseteq\mathcal V,\;
x\in\mathcal V\setminus B\\F(\{x\})>0}}
\frac{\Delta(x\mid B)}{F(\{x\})}.
\end{equation*}
\end{definition}
For every $B\subseteq\mathcal V$ and $x\in\mathcal V\setminus B$, $\Delta(x\mid B)\geq(1-c)F(\{x\})$, including when $F(\{x\})=0$. An additive function has $c = 0$. At the other extreme, $c = 1$ occurs exactly when some element with $F(\{x\}) > 0$ has zero marginal gain in the presence of a set of other elements.

\subsection{Submodular Multi-Agent MDP}
With submodularity and curvature defined, we now introduce the submodular multi-agent MDP on which our method builds.
\begin{definition}[Submodular multi-agent MDP]\label{def:submodular_mmdp}
A submodular multi-agent MDP is specified by the tuple
\begin{equation*}
\langle N, H, \mathcal{S}, \mathcal{A},
\{P_i\}_{i \in [N]}, \{\rho_i\}_{i \in [N]}, F\rangle,
\end{equation*}
where $N$ denotes the number of agents and $H$ is the horizon length. The finite state space of each agent is $\mathcal{S}$, also called the local state space. The joint state space is $\mathbf{S} = \mathcal{S}^N$. The finite action space of each agent is $\mathcal{A}$, and the joint action space is $\mathbf{A} = \mathcal{A}^N$. The transition dynamics of agent $i$ are $P_i=(P_{i,0},\ldots,P_{i,H-1})$, where $P_{i,t}: \mathcal{S} \times \mathcal{A} \rightarrow \mathcal{D}(\mathcal{S})$ and $P_{i,t}(\cdot \mid s, a)$ is the distribution of its state at time $t+1$. The transition dynamics are decoupled, so the joint transition dynamics $\mathbf P=(P_1,\ldots,P_N)$ give the joint next-state distribution
$$\mathbf{P}_{t}(\boldsymbol{s}' \mid \boldsymbol{s}, \boldsymbol{a}) = \prod_{i \in [N]} P_{i, t}(s'_{i} \mid s_{i}, a_{i}).$$
The initial state distribution of agent $i$ is $\rho_i \in \mathcal{D}(\mathcal{S})$. Initial states are independent, so $\boldsymbol{\rho} = \prod_{i \in [N]} \rho_i$. The common reward is a monotone submodular function $F: 2^{\mathcal{V}} \rightarrow \mathbb{R}_{\geq 0}$ \emph{over trajectories}. Its ground set $\mathcal V=\{0,\ldots,H\}\times\mathcal S$ contains all possible visits, with $\mathcal V_{t:t'}$ defined as in~\eqref{eq:restriction}. We assume that the reward function satisfies $F(B)>0$ for at least one $B\subseteq\mathcal V$.
\end{definition}

Agent $i$'s trajectory is the set of its visits,
$$\tau_i=\{(t,s_{i,t}):0\leq t\leq H\}\subseteq\mathcal V,$$
and the joint trajectory is $\boldsymbol\tau=(\tau_1,\ldots,\tau_N)$. In set operations we identify a tuple of trajectories with the union of its components. Using the notation in~\eqref{eq:restriction}, this gives
$$F(\boldsymbol\tau_{t:t'}):=F\Big(\bigcup_{i \in [N]}\tau_{i,t:t'}\Big).$$
For example, if $\tau_1=\{(0,s_1)\}$ and $\tau_2=\{(0,s_2)\}$, then $\boldsymbol\tau=(\tau_1,\tau_2)$ is treated as $\{(0,s_1),(0,s_2)\}$ whenever it appears in a set operation.

The formulation and analysis extend to monotone submodular rewards on the ground set $\{0,\ldots,H-1\}\times\mathcal S\times\mathcal A$ by including actions in the trajectories. For simplicity, we present the results using $\mathcal V=\{0,\ldots,H\}\times\mathcal S$.

\paragraph{Histories and policies.}
For a memory length $m\in\{0,\ldots,H\}$ and a time $t\in\{0,\ldots,H\}$, the $m$-step history of agent $i$ is
$$h^m_{i,t}=(s_{i,\max(t-m,0)},\ldots,s_{i,t-1})\in\mathcal H^m_t,$$
where $\mathcal H^m_t=\mathcal S^{\min(t,m)}$. It contains up to $m$ states of agent $i$ before time $t$. The joint $m$-step history is $\boldsymbol h^m_t=(h^m_{1,t},\ldots,h^m_{N,t})\in\boldsymbol{\mathcal H}^m_t$, where $\boldsymbol{\mathcal H}^m_t=(\mathcal H^m_t)^N$. For $m=H$, the current state $s_{i,t}$ and the history $h^H_{i,t}$ together form the full history $(s_{i,0},\ldots,s_{i,t})$.

For a memory of length $m$, a global policy of agent $i$ is a sequence $\pi_i=(\pi_{i, 0},\ldots,\pi_{i, H-1})$ with $\pi_{i, t}:\mathbf S\times\boldsymbol{\mathcal H}^m_t\to\mathcal D(\mathcal A)$, and $\Pi_m$ denotes the class of such policies. Here, $\pi_{i, t}(a\mid\boldsymbol s_t,\boldsymbol h^m_t)$ denotes the probability that agent $i$ takes action $a\in\mathcal A$ at time $t$ given $\boldsymbol s_t$ and $\boldsymbol h^m_t$. A local policy of agent $i$ instead depends only on its local state and $m$-step history $(s_{i,t},h^m_{i,t})$, with $\pi_{i,t}:\mathcal S\times\mathcal H^m_t\to\mathcal D(\mathcal A)$. We denote the class of such policies by $\Pi^{\mathrm{loc}}_m$. A joint policy is denoted by $\boldsymbol\pi=(\pi_1,\ldots,\pi_N)$, where $\pi_i$ is a global or local policy of agent $i$. The joint policy at time $t$ is $\boldsymbol\pi_t=(\pi_{1,t},\ldots,\pi_{N,t})$, and the actions of the agents are independent given $(\boldsymbol s_t,\boldsymbol h^m_t)$,
\begin{equation}\label{eq:product_policy}
\boldsymbol\pi_t(\boldsymbol a\mid\boldsymbol s_t,\boldsymbol h^m_t)=\prod_{i\in[N]}\pi_{i,t}(a_i\mid\boldsymbol s_t,\boldsymbol h^m_t).
\end{equation}
We use the notation $\boldsymbol\Pi_m$ and $\boldsymbol\Pi^{\mathrm{loc}}_m$ for the sets of joint global and joint local policies with memory $m$, respectively. The cases $m=0$ and $m=H$ give global Markov and global history-dependent policies.

\paragraph{Problem statement.}
Given a submodular multi-agent MDP, our goal is to find a joint policy $\boldsymbol{\pi} \in \boldsymbol{\Pi}_H$ that maximizes the expected reward:
\begin{equation}\label{eq:objective}
    J(\boldsymbol{\pi})
    =
    \mathbb{E}_{\boldsymbol\pi, \boldsymbol\rho, \mathbf P}
    \left[F(\boldsymbol{\tau})\right],
    \qquad
    \boldsymbol{\pi}^*
    \in
    \arg\max_{\boldsymbol{\pi} \in \boldsymbol{\Pi}_H} J(\boldsymbol{\pi}).
\end{equation}
The expectation is taken over joint trajectories $\boldsymbol\tau$ induced by the policy $\boldsymbol\pi$ under the initial distribution $\boldsymbol\rho$ and the transition dynamics $\mathbf P$. We abbreviate the expectation as $\mathbb E_{\boldsymbol\tau\sim\boldsymbol\pi}$ in the rest of the paper. Similarly, $\mathbb E_{\tau_i\sim\pi_i}$ denotes expectation under agent $i$'s local policy $\pi_i$, with $P_i$ and $\rho_i$ implicit. We write $J^*=J(\boldsymbol\pi^*)$ for the optimal value of~\eqref{eq:objective}. Restricting $\boldsymbol\pi^*$ to the product form~\eqref{eq:product_policy} does not reduce $J^*$, because an optimal policy of this form exists (Lemma~\ref{lem:optimal_product_policy}).

In this work, we assume known transition dynamics $\{P_i\}_{i\in[N]}$ and initial distributions $\{\rho_i\}_{i\in[N]}$. During policy optimization, we also assume that $F(B)$ can be evaluated exactly for every $B\subseteq\mathcal V$.

\begin{remark}\label{rem:special_cases}
For $N=1$, our formulation reduces to reinforcement learning with a submodular reward over the agent's trajectory~\citep{prajapat2024}. \citet{chen2026} instead consider a multi-agent setting, but their objective is a sum of monotone submodular stage rewards $f:2^{\mathcal S\times\mathcal A}\to\mathbb R_{\geq0}$, each evaluated on the state-action pairs of all agents at one time step. Their setting is the special case of our framework with ground set $\{0,\ldots,H-1\}\times\mathcal S\times\mathcal A$ and
$$F(B)=\sum_{t=0}^{H-1} f\big(\{(s,a):(t,s,a)\in B_t\}\big).$$
\end{remark}

\subsection{Traffic-Monitoring Example}
\label{subsec:traffic_example}
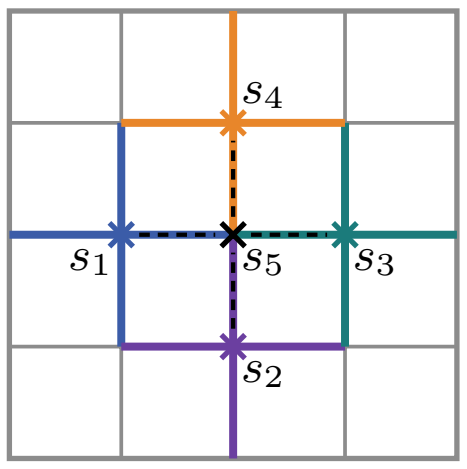
\begin{figure}[t]
\centering
\definecolor{gridline}{gray}{0.55}
\definecolor{scolA}{HTML}{3B5CA8}
\definecolor{scolB}{HTML}{6B3FA0}
\definecolor{scolC}{HTML}{1B7B7B}
\definecolor{scolD}{HTML}{E8862A}
\begin{tikzpicture}[scale=0.74]
  \draw[gridline, line width=0.7pt] (0,0) grid (4,4);
  \draw[gridline, line width=1pt] (0,0) rectangle (4,4);
  \foreach \col/\x/\y in {scolA/1/2,scolB/2/1,scolC/3/2,scolD/2/3}{
    \draw[\col,line width=1.5pt] (\x-1,\y)--(\x+1,\y) (\x,\y-1)--(\x,\y+1);
  }
  \foreach \ax/\ay/\bx/\by in {1.16/2/1.84/2,2.16/2/2.84/2,2/1.16/2/1.84,2/2.16/2/2.84}{
    \draw[black,line width=0.8pt,dash pattern=on 2pt off 1.5pt] (\ax,\ay)--(\bx,\by);
  }
  \foreach \col/\x/\y/\lab/\dx/\dy in {scolA/1/2/1/-0.272/-0.248,scolB/2/1/2/0.272/-0.248,scolC/3/2/3/0.272/-0.248,scolD/2/3/4/0.272/0.248,black/2/2/5/0.272/-0.248}{
    \draw[\col,line width=1pt] (\x-0.106,\y-0.106)--(\x+0.106,\y+0.106)
      (\x-0.106,\y+0.106)--(\x+0.106,\y-0.106);
    \node[font=\small] at ({\x+(\dx)},{\y+(\dy)}) {$s_{\lab}$};
  }
\end{tikzpicture}
\caption{Part of the road grid with five UAV positions $s_1,\ldots,s_5$. For $i\leq4$, the roads $\mathcal C(s_i)$ monitored from $s_i$ share the color of $s_i$. Each dashed road in $\mathcal C(s_5)$ is colored, so it is also monitored from one of $s_1,\ldots,s_4$.}
\label{fig:degenerate_curvature}
\end{figure}

We illustrate the submodular multi-agent MDP with a traffic-monitoring example used throughout the paper. In this example, the $N$ agents are moving UAVs that monitor an urban road network over a horizon of $H$ time steps. The network is an $M\times L$ grid with road set $\mathcal E$. The local state $s\in\mathcal S$ of each UAV is its position at a grid node, and its actions move it to an adjacent node or keep it in place. The transition dynamics and initial state distributions of the UAVs are decoupled, since the UAVs move independently and their initial positions are drawn independently. From position $s$, a UAV monitors the set $\mathcal C(s)\subseteq\mathcal E$ of roads connected to $s$. Figure~\ref{fig:degenerate_curvature} shows $\mathcal C(s)$ for five positions.

Monitoring each road $e\in\mathcal E$ provides information such as its traffic density and occupancy~\citep{maljkovic2025}. We call the added value of this information the \emph{importance} of the road. Before road $e$ is first monitored, its importance takes its maximum value $w_e\geq0$. When a UAV monitors a road $e$, the importance of the road drops to zero. Its importance then recovers over time as the information becomes outdated. After $\ell$ time steps, its importance is $w_e\alpha_\ell$, where $\alpha_\ell=1-e^{-\beta\ell}$ and $\beta>0$ is the recovery rate.

For a set $B\subseteq\mathcal V$ and a road $e\in\mathcal E$, the set of times at which road $e$ is monitored by visits in $B$ is
$$T_e(B)=\{t : (t,s)\in B \text{ for some } s \in \mathcal{S} \text{ with } e\in\mathcal C(s)\}.$$
For $t\in T_e(B)$, let $\ell_e(t)=t-\max\{t'\in T_e(B):t'<t\}$ be the number of time steps since road $e$ was previously monitored. If $t$ is the first time in $T_e(B)$, we set $\ell_e(t)=\infty$ and $\alpha_\infty=1$. The traffic-monitoring reward is
\begin{equation}\label{eq:traffic_reward}
  F(B)=\sum_{e\in\mathcal E}\,\sum_{t\in T_e(B)} w_e\,\alpha_{\ell_e(t)}.
\end{equation}
Thus, $F(\boldsymbol\tau)$ is the total value of the information collected by the fleet along its joint trajectory $\boldsymbol\tau$, and $F$ has diminishing returns across agents and over time. For example, when two UAVs monitor a road at the same time, the second UAV adds no new information about that road. To complete the formulation, we show in Appendix~\ref{app:traffic} that the reward~\eqref{eq:traffic_reward} is monotone submodular for a general class of recovery profiles $\alpha_\ell$, and this class includes $\alpha_\ell=1-e^{-\beta\ell}$ for every $\beta>0$.

\section{Hardness and Curvature}\label{sec:hardness_curvature}
To further motivate our results, we first explain why finding a policy that maximizes $J(\boldsymbol\pi)$ in~\eqref{eq:objective} is computationally hard. This hardness justifies restricting the class of rewards to obtain constant-factor guarantees. We then introduce pre-window curvature $c_m$, a property of the reward function that our approximation guarantees depend on.
\subsection{Computational Hardness}\label{subsec:intractability}
Maximizing $J$ poses two computational challenges. First, the single-agent case $N=1$ admits no constant-factor approximation in polynomial time under standard complexity assumptions, even with deterministic dynamics~\citep[Theorem~1]{prajapat2024}. For stochastic dynamics, optimal policies of this case may require history dependence, and representing an arbitrary history-dependent policy can require space exponential in the horizon~\citep{prajapat2024}. Second, the joint state and action spaces of $N$ agents have sizes $|\mathcal S|^N$ and $|\mathcal A|^N$, so optimizing directly over them takes time exponential in $N$.

Constant-factor guarantees therefore require further restricting the class of rewards. For a single agent, existing algorithms achieve a $(1-c)$-approximation for submodular trajectory rewards with total curvature $c<1$~\citep{prajapat2024,de2024global}. For multiple agents under decoupled dynamics, \citet{chen2026} obtain a $1/2$-approximation for time-additive rewards without any curvature assumption. Our framework contains both settings, so we seek a single curvature condition that yields an approximation guarantee for our problem and recovers or improves the known guarantees when specialized to either setting.

Total curvature does not achieve this goal in the multi-agent setting, since simultaneous visits can make $c=1$ even for time-additive rewards. Consider Figure~\ref{fig:degenerate_curvature} with the reward~\eqref{eq:traffic_reward} and $w_e>0$ for every road $e\in\mathcal E$. The visits $(t,s_1),\ldots,(t,s_4)$ monitor every road in $\mathcal C(s_5)$. Thus, $x=(t,s_5)$ has zero marginal gain given these four visits although $F(\{x\})=\sum_{e\in\mathcal C(s_5)}w_e>0$, so $c=1$ regardless of the choice of $\alpha_\ell$ in the reward~\eqref{eq:traffic_reward}. We therefore seek a condition that limits the effect of visits long before the current time on marginal gains.

\subsection{Pre-Window Curvature}\label{subsec:curvature}
We quantify how visits before an $m$-step window decrease the marginal gain of a current visit. For the ground set $\mathcal V$ of Definition~\ref{def:submodular_mmdp}, the $m$-step window at time $t$ is $\mathcal V_{t-m:t-1}\subseteq\mathcal V$ (Figure~\ref{fig:prewindow} in Appendix~\ref{app:curvature}). For fixed $x_t\in\mathcal V_t$ and $Z\subseteq\mathcal V$, adding visits $B\subseteq\mathcal V_{0:t-m-1}$ to $Z$ changes the marginal gain from $\Delta(x_t\mid Z)$ to $\Delta(x_t\mid B\cup Z)$. We measure this decrease relative to $\Delta(x_t\mid Z)$ and exclude cases with $\Delta(x_t\mid Z)=0$, since monotonicity and submodularity then give $\Delta(x_t\mid B\cup Z)=0$. We define pre-window curvature as the largest relative decrease over all such $x_t$, $Z$, and $B$.
\begin{definition}[Pre-window curvature]\label{def:windowed_curvature}
For $m\in\{0,\ldots,H\}$, the pre-window curvature of $F$ is
\begin{equation}\label{eq:window_curvature}
 c_m=1-\min_{\substack{t,\;x_t,\;B,\;Z\\ \Delta(x_t\mid Z)>0}}
 \frac{\Delta(x_t\mid B\cup Z)}{\Delta(x_t\mid Z)},
\end{equation}
where the minimum ranges over times $t\in\{0,\ldots,H\}$, visits $x_t\in\mathcal V_t$, and sets $B\subseteq\mathcal V_{0:t-m-1}$ and $Z\subseteq\mathcal V$.
\end{definition}
Pre-window curvature satisfies $0\leq c_m\leq c$ and is nonincreasing in $m$, with $c_H=0$. Moreover, $c_m=0$ whenever $F$ is additive over time, even if overlap across agents makes $c$ equal one. We prove these properties in Appendix~\ref{app:curvature}.

Returning to the example of Section~\ref{subsec:traffic_example}, we show that the reward~\eqref{eq:traffic_reward} has pre-window curvature $c_m<1$ for every $\beta>0$. Proposition~\ref{prop:traffic_curvature} in Appendix~\ref{subsec:app_exp} gives the explicit expression $c_m=e^{-\beta(m+1)}$ for $0\leq m<H$. Without recovery ($\beta=0$), the reward~\eqref{eq:traffic_reward} values only the first monitoring of each road and has $c_m=c=1$ for every $m<H$. In summary, $c_m<1$ describes rewards in which visits before an $m$-step window have a limited effect on the marginal gain of a current visit.

\input{method}

\begin{figure*}[t]
\centering
\setlength{\tabcolsep}{14pt}
\begin{tabular}{@{}cc@{}}
\includegraphics{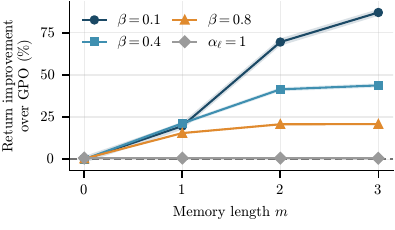} &
\includegraphics{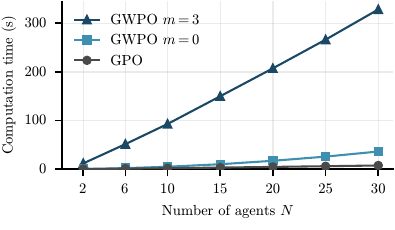} \\
{\footnotesize (a) Value of memory} &
{\footnotesize (b) Computation time} \\
\end{tabular}
\caption{Performance improvement and computational cost of memory in GWPO. (a)~Return improvement of GWPO with different memory lengths $m$ over \textsc{GPO}~\citep{chen2026} for $N=20$ agents and different recovery rules. (b)~Computation time for different numbers of agents $N$, comparing GWPO with $m=0,3$ and \textsc{GPO} at $\beta=0.4$.}
\label{fig:gpo_comparison}
\end{figure*}

\begin{figure*}[t]
\centering
\setlength{\tabcolsep}{10pt}
\begin{tabular}{@{}ccc@{}}
\includegraphics{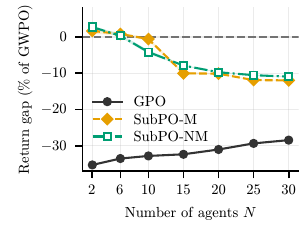} &
\includegraphics{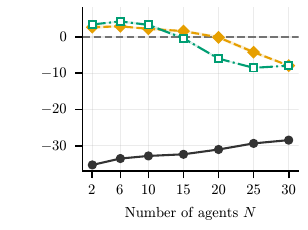} &
\includegraphics{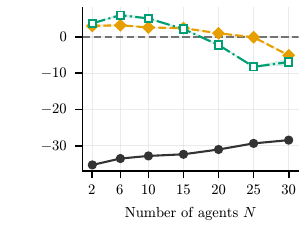} \\
{\footnotesize (a) Computation time $T_N$} &
{\footnotesize (b) Computation time $5T_N$} &
{\footnotesize (c) Computation time $20T_N$} \\
\end{tabular}
\caption{Return gap between each baseline and GWPO with $m=3$ at $\beta=0.4$, as a percentage of the GWPO return. Negative values indicate higher returns for GWPO. \textsc{SubPO-M} and \textsc{SubPO-NM} are allowed computation times of (a)~$T_N$, (b)~$5T_N$, and (c)~$20T_N$, where $T_N$ is the computation time of GWPO for $N$ agents.}
\label{fig:budget_comparison}
\end{figure*}

\section{Numerical Experiments}\label{sec:experiments}
We empirically studied the return and computation time of GWPO with memory length $m$ on the traffic-monitoring example of Section~\ref{subsec:traffic_example}. Our instance used a $20\times20$ road grid, the horizon $H=20$, and initial positions of the agents sampled independently and uniformly. We first studied how memory affected the return and the computation time of GWPO, and then compared its performance with two baselines. The baselines were Greedy Policy Optimization (\textsc{GPO})~\citep{chen2026} and Submodular Policy Optimization (\textsc{SubPO})~\citep{prajapat2024}. \textsc{GPO} optimizes a time-additive objective, and \textsc{SubPO} optimizes global policies by policy gradient. We used \textsc{SubPO} with Markov (\textsc{SubPO-M}) and history-dependent (\textsc{SubPO-NM}) policies. \textsc{SubPO-NM} encodes the history with a recurrent neural network. We evaluated all policies under the reward~\eqref{eq:traffic_reward} on $10{,}000$ common initial joint states. Appendix~\ref{app:experimental_details} gives further experimental details.

Figure~\ref{fig:gpo_comparison} shows how the memory length affects the return and computation time of GWPO. For $N=20$ agents, increasing memory improved returns over \textsc{GPO} for the non-additive rewards in Figure~\ref{fig:gpo_comparison}(a). For the recovery rule $\alpha_\ell=1-e^{-\beta\ell}$, smaller $\beta$ means that road importance recovers more slowly after a visit. Longer memory helped more when recovery was slower, while GWPO approximately matched \textsc{GPO} in the time-additive case $\alpha_\ell=1$. Figure~\ref{fig:gpo_comparison}(b) shows that these gains came with higher computation time.

Figure~\ref{fig:budget_comparison} compares return gaps relative to GWPO across the number of agents and computation times. With $T_N$, GWPO and \textsc{SubPO} achieved comparable returns for $N\leq6$, while GWPO achieved higher returns for $N\geq10$. With $20T_N$, both \textsc{SubPO} variants achieved up to $6\%$ higher returns than GWPO for $N\leq15$, but lower returns at $N=30$. Appendix~\ref{app:experiment_results} reports results over five \textsc{SubPO} seeds and their confidence intervals.

\section{Conclusion}
We introduced pre-window curvature to quantify how visits before an $m$-step window affect marginal gains in submodular rewards over trajectories. Using this notion, we developed GWPO, which exploits decoupled dynamics to sequentially optimize local policies. For fixed $m$, its computation time is polynomial in the number of agents and the horizon. We established a $\frac{1-c_m}{2-c_m}$ approximation guarantee relative to the optimal global history-dependent return, up to estimation error. We formulated multi-agent traffic monitoring as a submodular multi-agent MDP and evaluated GWPO against the time-additive baseline \textsc{GPO} and \textsc{SubPO}, which optimizes global policies by policy gradient.

\subsubsection*{Acknowledgements}
This work was supported as a part of NCCR Automation, a National Centre of Competence in Research, funded by the Swiss National Science Foundation (grant number 51NF40\_225155).

\subsection*{AI use statement}
In this work, we used generative AI tools for providing feedback on experimental design, checking proofs and suggesting minor corrections, and assisting with implementing and editing code for numerical experiments.
We have not used generative AI tools for developing theoretical models or conceptual frameworks, formulating mathematical claims, providing critical ingredients for proofs, proposing or refining hypotheses, or interpreting results.
Generating synthetic datasets, cleaning or reformatting datasets, qualitative or thematic data analysis, and translation are not applicable to this work.
Additionally, we used generative AI tools for creating and modifying figures, editing text to improve readability, and finding additional related work.
We have reviewed all AI outputs used in the activities mentioned above. LLM-generated code was verified and tested for correctness by the authors.  All related work identified using AI tools was read and checked by the authors. The authors checked all AI-edited text and AI-suggested revisions.
We take responsibility for the final content of this work, including text, claims, or artifacts produced with the aid of generative AI.

\bibliographystyle{plainnat}
\bibliography{references}

\clearpage
\appendix
\thispagestyle{fancy}

\onecolumn
\aistatstitle{Appendix}

\input{appendix}
\end{document}

%% file: method.tex
\section{Algorithms and Analysis}\label{sec:method}
To address the challenges of Section~\ref{subsec:intractability}, we first approximate the objective $J$ by the expected sum of stage rewards that depend only on the current joint state and the joint $m$-step history. The error of this approximation is controlled by the pre-window curvature $c_m$, and the maximizer of this approximated objective achieves at least $(1-c_m)J^*$. However, maximizing this objective over joint policies is still exponential in the number of agents $N$. In Section~\ref{subsec:sequential}, we therefore exploit the decoupled dynamics to optimize the policies of the agents sequentially in polynomial time for fixed $m$. The resulting policies achieve a $\frac{1-c_m}{2-c_m}$ guarantee up to estimation error.

\subsection{History-Augmented State}\label{subsec:augmented}
We introduce the history-augmented state used to define the finite-horizon MDPs in the next two subsections. We fix a memory length $m\in\{0,\ldots,H\}$ and write $h_{i,t}=h^m_{i,t}$ for the histories of Section~\ref{sec:problem}. At time $t$, the history-augmented state of agent $i$ is the pair $\tilde s_{i,t}=(s_{i,t},h_{i,t})$, which lies in the augmented state space $\tilde{\mathcal S}_t=\mathcal S\times\mathcal H^m_t$. By definition, policies in $\Pi^{\mathrm{loc}}_m$ and $\boldsymbol\Pi_m$ are Markov policies on $\tilde s_{i,t}$ and $\tilde{\boldsymbol{s}}_t\in\tilde{\mathbf{S}}_t=\tilde{\mathcal S}_t^N$, respectively. For $\pi_i\in\Pi^{\mathrm{loc}}_m$, the input of $\pi_{i,t}$ takes $|\tilde{\mathcal S}_t|=|\mathcal S|^{\min(t,m)+1}\leq|\mathcal S|^{m+1}$ values, compared with $|\mathcal S|^{t+1}$ values for a full history.

The augmented transition dynamics are Markov and remain decoupled across agents. In particular, the next state $s_{i,t+1}\sim P_{i,t}(\cdot\mid s_{i,t},a_{i,t})$ depends only on the state and action of agent $i$. The history updates deterministically by appending $s_{i,t}$ to $h_{i,t}$ and dropping the oldest entry if its length exceeds $m$. Figure~\ref{fig:augmented_state} in Appendix~\ref{app:augmented_dp} illustrates this update.

Since the reward function $F$ is defined on sets of visits, we map each history to the visits it determines. The map $\nu_t:\mathcal H^m_t\to2^{\mathcal V}$ assigns to each history $h=(s_{\max(t-m,0)},\ldots,s_{t-1})\in\mathcal H^m_t$ the set of visits
\begin{equation*}
 \nu_t(h)=\big\{(l,s_l):\max(t-m,0)\leq l\leq t-1\big\}.
\end{equation*}
For a joint history, $\nu_t(\boldsymbol h_t)=\bigcup_{i\in[N]}\nu_t(h_{i,t})$. Along a trajectory, $\nu_t(h_{i,t})=\tau_{i,t-m:t-1}$ and $\nu_t(\boldsymbol h_t)=\boldsymbol\tau_{t-m:t-1}$.

\subsection{Windowed Objective}\label{subsec:surrogate}
Our goal is to approximate Problem~\eqref{eq:objective} by a finite-horizon MDP with additive stage rewards on the joint augmented states. We therefore need stage rewards $r_t(\tilde{\boldsymbol s}_t)$ that depend only on the joint augmented state and whose expected sum under each policy $\boldsymbol\pi$ approximates $J(\boldsymbol\pi)$. Adding the team visits one time step at a time decomposes $F$ over time as
\begin{equation}\label{eq:F_telescope}
 F(\boldsymbol\tau)=\sum_{t=0}^{H}
 \Delta(\boldsymbol\tau_t\mid\boldsymbol\tau_{0:t-1}).
\end{equation}
The term at time $t$ is the marginal gain of the current team visits $\boldsymbol\tau_t$ given all earlier visits $\boldsymbol\tau_{0:t-1}$. The augmented state $\tilde{\boldsymbol s}_t$ determines only the part $\boldsymbol\tau_{t-m:t-1}$ of these earlier visits, so this term is not always a stage reward on the augmented states. We therefore define the stage reward $r_t:\tilde{\mathbf S}_t\to\mathbb R_{\geq0}$ as the marginal gain of the current team visits given only the visits $\nu_t(\boldsymbol h_t)$ within the $m$-step window,
\begin{equation}\label{eq:stage_reward}
 r_t(\tilde{\boldsymbol{s}}_t)
 =\Delta\big(\{(t,s_{i,t}):i\in[N]\}\mid\nu_t(\boldsymbol h_t)\big).
\end{equation}
Along a trajectory, $r_t(\tilde{\boldsymbol s}_t)=\Delta(\boldsymbol\tau_t\mid\boldsymbol\tau_{t-m:t-1})$. Replacing each term of Equation~\eqref{eq:F_telescope} by $r_t(\tilde{\boldsymbol s}_t)$ and taking the expectation gives the \emph{windowed objective},
\begin{equation*}
 \Phi_m(\boldsymbol\pi)
 =\mathbb E_{\boldsymbol\tau\sim\boldsymbol\pi}
 \left[\sum_{t=0}^H r_t(\tilde{\boldsymbol s}_t)\right].
\end{equation*}
Maximizing $\Phi_m$ is thus a finite-horizon MDP problem on the augmented states. This problem has a deterministic optimal policy that lies in $\boldsymbol\Pi_m$~\citep{puterman1994}. Algorithm~\ref{alg:augmented_dp} in Appendix~\ref{app:augmented_dp} computes such a policy. It remains to show that $\Phi_m$ approximates $J$.

\begin{restatable}{lemma}{WindowedSandwich}\label{lem:sandwich}
For every policy $\boldsymbol\pi$,
\begin{equation*}
 (1-c_m)\,\Phi_m(\boldsymbol\pi)\leq J(\boldsymbol\pi)\leq\Phi_m(\boldsymbol\pi).
\end{equation*}
\end{restatable}
\emph{Proof sketch.} The upper bound holds by submodularity, since $r_t$ considers only $\boldsymbol\tau_{t-m:t-1}\subseteq\boldsymbol\tau_{0:t-1}$. The lower bound holds by the definition of $c_m$.

\begin{restatable}{theorem}{SurrogateGuarantee}\label{thm:surrogate_guarantee}
Let $\boldsymbol{\pi}'\in\arg\max_{\boldsymbol\pi\in\boldsymbol\Pi_H}\Phi_m(\boldsymbol\pi)$. Then
\begin{equation*}
J(\boldsymbol{\pi}') \geq (1-c_m)\,J^*.
\end{equation*}
\end{restatable}
\emph{Proof sketch.} The guarantee follows from Lemma~\ref{lem:sandwich} and the optimality of $\boldsymbol\pi'$ for $\Phi_m$. We prove both results in Appendix~\ref{app:windowed_sandwich}.

For a single agent with $m=0$, the windowed objective is the expected modular reward $\sum_{x\in\boldsymbol\tau}F(\{x\})$ of~\citet[Proposition~3]{prajapat2024}, and our proof follows their argument with the factor $1-c_0\geq1-c$. For $m=H$, $c_H=0$ and $\Phi_H=J$, so the theorem yields an optimal policy. However, the MDP has up to $|\mathcal S|^{N(m+1)}$ joint augmented states per time step and $|\mathcal A|^N$ joint actions. This exponential dependence on $N$ motivates the sequential policy optimization introduced next.

\subsection{Sequential Policy Optimization}\label{subsec:sequential}
We adopt the sequential policy optimization approach of prior work~\citep{singh2009,chen2026} to avoid joint optimization. This approach optimizes the policy of one agent at a time while holding predecessor policies fixed. However, the guarantee of~\citet{chen2026} cannot be applied directly to $\Phi_m$. The reason is that an added agent's visits within the window can reduce current marginal gains, which violates the required monotonicity assumption. We illustrate this violation with an example in Appendix~\ref{app:windowed_nonmonotonicity}.

Instead, we sequentially construct local stage rewards using complete predecessor trajectories and each agent's own $m$-step history. We then use pre-window curvature $c_m$ to bound the relative overestimate caused by ignoring the agent's visits before the window, which allows a guarantee relative to the optimal global history-dependent return $J^*$.

We optimize the policies of the agents in the order $1,\ldots,N$. For agent $i$, we fix the predecessor policies $\bar{\boldsymbol\pi}_{[i-1]}=(\bar\pi_1,\ldots,\bar\pi_{i-1})$ computed so far. Even with fixed policies, predecessor trajectories $\boldsymbol\tau_{[i-1]}$ can be random because of stochastic dynamics or random initial states. We therefore take the expectation over $\boldsymbol\tau_{[i-1]}\sim\bar{\boldsymbol\pi}_{[i-1]}$ to define the local stage reward $r_{i,t}^{\bar{\boldsymbol\pi}_{[i-1]}}:\tilde{\mathcal S}_t\to\mathbb R_{\geq0}$ as
\begin{equation}\label{eq:local_reward}
 r_{i,t}^{\bar{\boldsymbol\pi}_{[i-1]}}(\tilde s_{i,t})=
 \mathbb E\!\left[
 \Delta\big((t,s_{i,t})\mid
 \nu_t(h_{i,t})\cup\boldsymbol\tau_{[i-1]}\big)
 \right].
\end{equation}
For $i=1$, the reward $r_{1,t}^{\bar{\boldsymbol\pi}_{[0]}}(\tilde s_{1,t})=\Delta\big((t,s_{1,t})\mid\nu_t(h_{1,t})\big)$ is the stage reward~\eqref{eq:stage_reward} with $N=1$. For a policy $\pi_i$ of agent $i$, the local objective is the expected sum of the rewards $r_{i,t}^{\bar{\boldsymbol\pi}_{[i-1]}}$ along the trajectory $\tau_i$ induced by $\pi_i$,
\begin{equation}\label{eq:G_objective}
 G_i^{\bar{\boldsymbol\pi}_{[i-1]}}(\pi_i)
 =
 \mathbb E_{\tau_i\sim\pi_i}\left[
 \sum_{t=0}^H
 r_{i,t}^{\bar{\boldsymbol\pi}_{[i-1]}}(\tilde s_{i,t})
 \right].
\end{equation}
As in Section~\ref{subsec:surrogate}, maximizing $G_i^{\bar{\boldsymbol\pi}_{[i-1]}}$ is an MDP problem on the augmented states of agent $i$. This problem has a deterministic optimal policy that lies in $\Pi^{\mathrm{loc}}_m$.

The exact evaluation of~\eqref{eq:local_reward} may require summing over up to $|\mathcal S|^{(i-1)(H+1)}$ predecessor trajectories. For each agent $i\geq2$, we therefore estimate this expectation using $K$ trajectories $\boldsymbol\tau_{[i-1]}^1,\ldots,\boldsymbol\tau_{[i-1]}^K$ sampled independently under $\bar{\boldsymbol\pi}_{[i-1]}$,
\begin{equation}\label{eq:empirical_reward}
 \widehat r_{i,t}^{\bar{\boldsymbol\pi}_{[i-1]}}(\tilde s_{i,t})
 =
 \frac{1}{K}\sum_{k=1}^{K}
 \Delta\big((t,s_{i,t})\mid
 \nu_t(h_{i,t})\cup\boldsymbol\tau_{[i-1]}^{k}\big).
\end{equation}
Replacing $r_{i,t}^{\bar{\boldsymbol\pi}_{[i-1]}}$ with $\widehat r_{i,t}^{\bar{\boldsymbol\pi}_{[i-1]}}$ in~\eqref{eq:G_objective} defines the empirical local objective $\widehat G_i^{\bar{\boldsymbol\pi}_{[i-1]}}$. For agent $1$, no sampling is required and $\widehat G_1=G_1$. 

Greedy Windowed Policy Optimization (GWPO), stated in Algorithm~\ref{alg:GWPO}, solves each agent's augmented MDP in turn with any exact finite-horizon MDP solver, such as Algorithm~\ref{alg:augmented_dp}. For agent $i$, it maximizes $\widehat G_i^{\bar{\boldsymbol\pi}_{[i-1]}}$ given the fixed predecessors' policies $\bar{\boldsymbol\pi}_{[i-1]}$, and then fixes $\bar\pi_i$ before proceeding to agent $i+1$. During centralized training, GWPO samples the trajectories of the predecessors and evaluates $F$ exactly, as assumed in Section~\ref{sec:problem}. Execution is decentralized, since each agent's policy depends only on its own augmented state.

\begin{algorithm}[t]
\caption{GWPO}
\label{alg:GWPO}
\begin{algorithmic}[1]
\REQUIRE submodular multi-agent MDP; memory length $m\in\{0,\ldots,H\}$; number of samples $K\geq1$
\FOR{$i \in [N]$}
  \IF{$i=1$}
    \STATE $\widehat r_{1,t}^{\bar{\boldsymbol\pi}_{[0]}}(\tilde s_{1,t})
      =\Delta\big((t,s_{1,t})\mid\nu_t(h_{1,t})\big)$
      \quad $\forall t,\tilde s_{1,t}$
  \ELSE
    \STATE $\boldsymbol\tau_{[i-1]}^{1},\ldots,
      \boldsymbol\tau_{[i-1]}^{K}\sim\bar{\boldsymbol\pi}_{[i-1]}$
    \STATE Estimate $r_{i,t}^{\bar{\boldsymbol\pi}_{[i-1]}}(\tilde s)$ by $\widehat r_{i,t}^{\bar{\boldsymbol\pi}_{[i-1]}}(\tilde s)$ in~\eqref{eq:empirical_reward}\enspace $\forall t,\tilde s$
  \ENDIF
  \STATE $\bar\pi_i\leftarrow$
    Algorithm~\ref{alg:augmented_dp}$\big(\mathcal S,\mathcal A,m,
    P_i,\widehat r_{i,t}^{\bar{\boldsymbol\pi}_{[i-1]}}\big)$
\ENDFOR
\RETURN $\bar{\boldsymbol\pi}=(\bar\pi_1,\ldots,\bar\pi_N)$
\end{algorithmic}
\end{algorithm}

The guarantee of GWPO depends on the number of agents. For $N=1$, GWPO exactly maximizes $\Phi_m$, so Theorem~\ref{thm:surrogate_guarantee} gives $J(\bar{\boldsymbol\pi})\geq(1-c_m)J^*$. For $N\geq2$, the following theorem bounds $J(\bar{\boldsymbol\pi})$ relative to $J^*$ in terms of $c_m$ and the reward-estimation error.
\begin{restatable}[]{theorem}{SequentialGuarantee}\label{thm:sequential_guarantee}
Let $N\geq2$ and fix $m\in\{0,\ldots,H\}$. For $\epsilon>0$ and $\delta\in(0,1)$, choose the number of sampled predecessor trajectories $K$ such that
\begin{equation}\label{eq:sample_complexity}
 K\geq \frac{R^2}{2\epsilon^2}
 \ln\frac{2(N-1)(H+1)|\mathcal S|^{m+1}}{\delta},
\end{equation}
where $R=\max_{v\in\mathcal V}F(\{v\})$ is the largest reward of a single visit. Then, with probability at least $1-\delta$, the joint policy $\bar{\boldsymbol\pi}$ returned by Algorithm~\ref{alg:GWPO} satisfies
\begin{equation}\label{eq:product_guarantee}
 J(\bar{\boldsymbol\pi})
 \geq\frac{1-c_m}{2-c_m}
 \left(J^*-2(N-1)(H+1)\epsilon\right).
\end{equation}
\end{restatable}

The approximation factor $\frac{1-c_m}{2-c_m}$ is positive for $c_m<1$, equals $\frac12$ at $c_m=0$, and is zero at $c_m=1$. The additive error comes from GWPO optimizing the estimated rewards $\widehat r_{i,t}^{\bar{\boldsymbol\pi}_{[i-1]}}$ instead of the rewards $r_{i,t}^{\bar{\boldsymbol\pi}_{[i-1]}}$ for each agent $i\geq2$ and time $t$. By~\eqref{eq:sample_complexity}, the sample size $K$ grows only logarithmically in $N$, $H$, and $1/\delta$. 

For rewards additive over time, Proposition~\ref{prop:cm_properties}(iv) shows that $c_m=0$ for every $m$. Therefore, the factor equals $\frac12$, as in~\citet[Theorem~1]{chen2026}. The factor $\frac{1-c_m}{2-c_m}$ has the form $\frac{q}{1+q}$ of sequential allocation~\citep[Theorem~4]{singh2009adaptive} with $q=1-c_m$. Their theorem assumes a $q$-approximation for the marginal reward of each agent and uses the same policy classes for the returned policies and the optimum. We do not assume such an approximation, and we compare our local policies with the optimal global history-dependent policy.

\begin{proof}[Proof sketch of Theorem~\ref{thm:sequential_guarantee}]
The proof addresses two challenges that do not arise in the time-additive setting~\citep{chen2026}. First, the optimal policy $\boldsymbol\pi^*\in\boldsymbol\Pi_H$ may depend on the full joint history, while the policy $\bar{\boldsymbol\pi}\in\boldsymbol\Pi^{\mathrm{loc}}_m$ returned by GWPO uses only the current state and $m$-step history of each agent. Second, the return $J(\bar{\boldsymbol\pi})$ is not the sum of the objectives that GWPO maximizes, since the objective of each agent ignores its visits prior to the $m$-step window. To address the first challenge, we consider an arbitrary joint policy $\boldsymbol\pi \in \boldsymbol\Pi_H$ and an arbitrary agent $i$. We show that there exists a local policy in $\Pi^{\mathrm{loc}}_m$ that reproduces the distribution of the augmented state of agent $i$ under $\boldsymbol\pi$ at every time step. For the second, we use pre-window curvature to lower bound $J(\bar{\boldsymbol\pi})$ by the objectives that GWPO maximizes.

The proof proceeds in three steps. First, the optimality gap $J^*-J(\bar{\boldsymbol\pi})$ is bounded by the sum of the maximum local objectives of the agents. Second, we bound the suboptimality of maximizing the local objective of each agent by the errors of reward estimation. Following~\citet{chen2026}, Hoeffding's inequality and a union bound keep all these errors below $\epsilon$ with probability at least $1-\delta$. Finally, pre-window curvature lower bounds $J(\bar{\boldsymbol\pi})$ by $1-c_m$ times the sum of the attained local objectives. Together, these three bounds imply~\eqref{eq:product_guarantee}. Appendix~\ref{app:sequential} gives the full proof.
\end{proof}

GWPO achieves the guarantee of Theorem~\ref{thm:sequential_guarantee} in polynomial time for fixed $m$. We count oracle evaluations and transition samples as unit-cost operations and first consider the cost for a single agent $i$. Sampling $K$ trajectories of its at most $N-1$ predecessors costs $O(NKH)$. Estimating the rewards~\eqref{eq:empirical_reward} at all times and augmented states costs $O(H|\mathcal S|^{m+1}K)$. Solving the MDP of agent $i$ on its augmented states with Algorithm~\ref{alg:augmented_dp} costs $O(H|\mathcal S|^{m+2}|\mathcal A|)$. Summing over the $N$ agents gives the computation time of Algorithm~\ref{alg:GWPO},
\begin{equation*}
 O\!\left(
 NH|\mathcal S|^{m+1}(K+|\mathcal S||\mathcal A|)
 +N^2KH
 \right).
\end{equation*}

%% file: appendix.tex
\section{Correlated Policies}\label{app:correlated}
The class $\boldsymbol\Pi_H$ uses the product action distributions in~\eqref{eq:product_policy} with $m=H$. We show that allowing joint action distributions that do not factor in this way does not improve the optimal expected reward.
The following lemma applies the standard existence result for deterministic optimal policies in finite-horizon MDPs~\citep[Proposition~4.4.3]{puterman1994} to the augmented state $(\boldsymbol s_t,\boldsymbol h_t^H)$.
\begin{lemma}
\label{lem:optimal_product_policy}
There exists a deterministic policy $\boldsymbol\pi^*\in\boldsymbol\Pi_H$ that is optimal among all global history-dependent joint policies, including those with correlated actions. Consequently, allowing correlated action distributions does not increase $J^*$.
\end{lemma}

\begin{proof}
At each time $t\in\{0,\ldots,H-1\}$, allow an arbitrary joint action distribution
$$
q_t:\mathbf S\times\boldsymbol{\mathcal H}_t^H\to\mathcal D(\mathbf A).
$$
For every joint history $(\boldsymbol s_t,\boldsymbol h_t^H)$, this means that
$$
q_t(\boldsymbol a\mid\boldsymbol s_t,\boldsymbol h_t^H)\geq0
\quad\text{for all }\boldsymbol a\in\mathbf A,
\qquad
\sum_{\boldsymbol a\in\mathbf A}q_t(\boldsymbol a\mid\boldsymbol s_t,\boldsymbol h_t^H)=1.
$$
The joint action distribution $q_t$ does not necessarily factor across agents as in~\eqref{eq:product_policy}. Taking $(\boldsymbol s_t,\boldsymbol h_t^H)$ as the state yields a finite-horizon MDP. The transition dynamics of this MDP draw the next joint state from $\mathbf P_t(\cdot\mid\boldsymbol s_t,\boldsymbol a_t)$ and update the history by appending $\boldsymbol s_t$. Its reward is zero at intermediate stages and equals $F(\boldsymbol\tau)$ at time $H$, when the augmented state determines the full joint trajectory.

Fix optimal policies at time steps $t+1,\ldots,H-1$. At each joint state and history, the expected return is linear in $q_t(\cdot\mid\boldsymbol s_t,\boldsymbol h_t^H)$. Its maximum over the probability simplex $\mathcal D(\mathbf A)$ is therefore attained at a vertex, which assigns probability one to a single joint action. Backward induction thus yields a deterministic joint policy that is optimal under any initial distribution, including $\boldsymbol\rho$.

Let $\boldsymbol a_t^*(\boldsymbol s_t,\boldsymbol h_t^H)$ denote the joint action selected by backward induction, and let $a_{i,t}^*(\boldsymbol s_t,\boldsymbol h_t^H)$ denote its $i$-th component. For each agent $i\in[N]$, define
$$
\pi_{i,t}^*(a_i\mid\boldsymbol s_t,\boldsymbol h_t^H)
=
\mathbf 1\!\left\{
a_i=a_{i,t}^*(\boldsymbol s_t,\boldsymbol h_t^H)
\right\}.
$$
The product of these policies assigns probability one to the selected joint action. Hence, this optimal policy belongs to $\boldsymbol\Pi_H$, and allowing correlated action distributions does not increase $J^*$.
\end{proof}

\section{Properties of Pre-Window Curvature}\label{app:curvature}
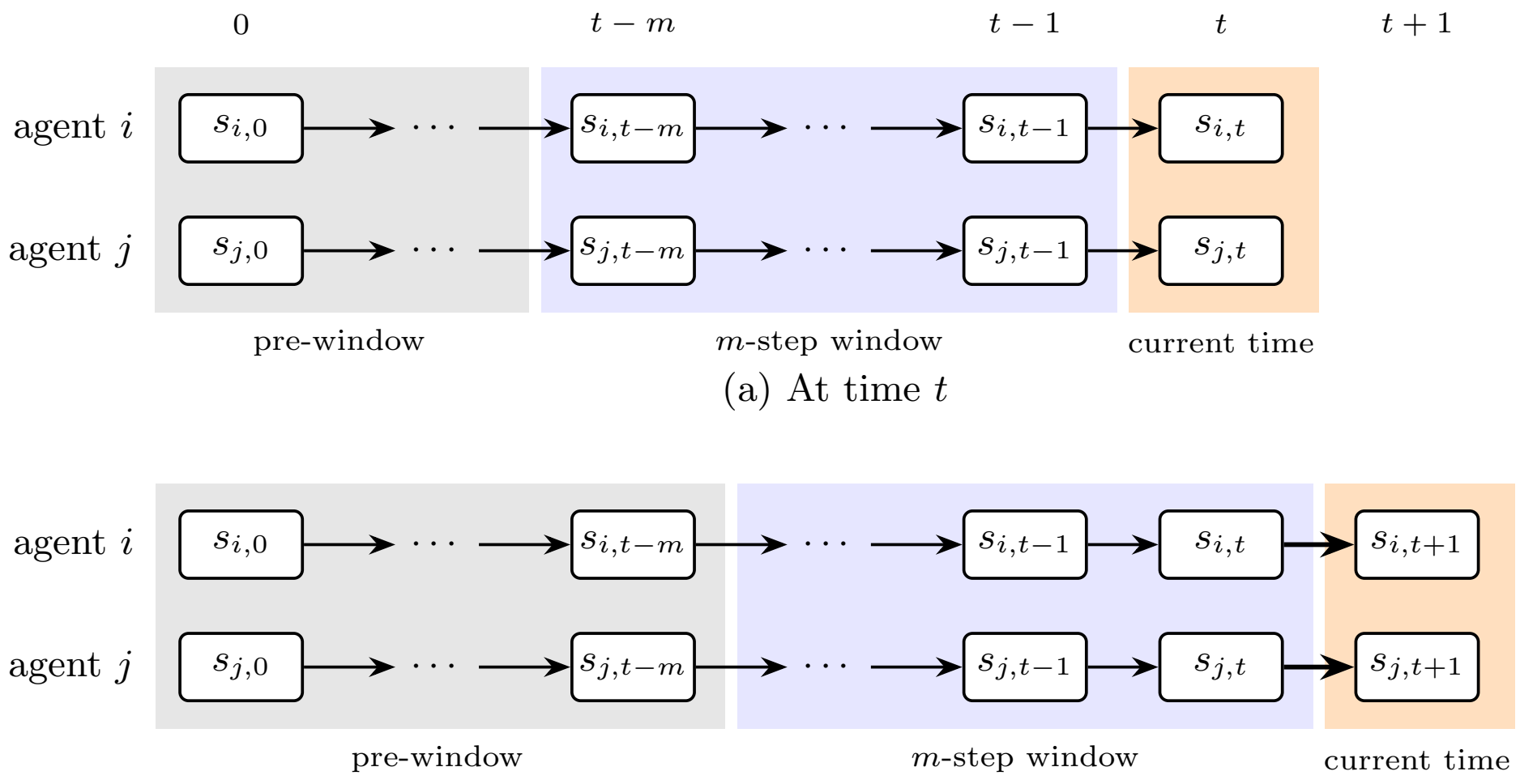
\begin{figure}[t]
\centering
\begin{tikzpicture}[font=\footnotesize,
  st/.style={draw, rounded corners=2pt, minimum width=1cm, minimum height=0.55cm,
             inner sep=1pt, line width=0.7pt, fill=white},
  tr/.style={-{Stealth}, line width=0.7pt}]
  \foreach \x/\lab in {0/0, 3.2/t-m, 6.4/t-1, 8.0/t, 9.6/t+1}{
    \node[font=\scriptsize] at (\x,0.85) {$\lab$};}
  \fill[black!10]  (-0.7,-1.5) rectangle (2.35,0.5);
  \fill[blue!10]   (2.45,-1.5) rectangle (7.15,0.5);
  \fill[orange!25] (7.25,-1.5) rectangle (8.8,0.5);
  \node[font=\scriptsize] at (0.8,-1.75) {pre-window};
  \node[font=\scriptsize] at (4.8,-1.75) {$m$-step window};
  \node[font=\scriptsize] at (8.0,-1.75) {current time};
  \foreach \r/\y in {i/0, j/-1}{
    \node[st] (a\r0) at (0,\y)   {$s_{\r,0}$};
    \node     (a\r1) at (1.6,\y) {$\cdots$};
    \node[st] (a\r2) at (3.2,\y) {$s_{\r,t-m}$};
    \node     (a\r3) at (4.8,\y) {$\cdots$};
    \node[st] (a\r4) at (6.4,\y) {$s_{\r,t-1}$};
    \node[st] (a\r5) at (8.0,\y) {$s_{\r,t}$};
    \foreach \p/\q in {0/1,1/2,2/3,3/4,4/5}{\draw[tr] (a\r\p) -- (a\r\q);}
    \node[anchor=east] at (-0.75,\y) {agent $\r$};
  }
  \node at (4.85,-2.15) {(a) At time $t$};
  \fill[black!10]  (-0.7,-4.9) rectangle (3.95,-2.9);
  \fill[blue!10]   (4.05,-4.9) rectangle (8.75,-2.9);
  \fill[orange!25] (8.85,-4.9) rectangle (10.4,-2.9);
  \node[font=\scriptsize] at (1.6,-5.15) {pre-window};
  \node[font=\scriptsize] at (6.4,-5.15) {$m$-step window};
  \node[font=\scriptsize] at (9.6,-5.15) {current time};
  \foreach \r/\y in {i/-3.4, j/-4.4}{
    \node[st] (b\r0) at (0,\y)   {$s_{\r,0}$};
    \node     (b\r1) at (1.6,\y) {$\cdots$};
    \node[st] (b\r2) at (3.2,\y) {$s_{\r,t-m}$};
    \node     (b\r3) at (4.8,\y) {$\cdots$};
    \node[st] (b\r4) at (6.4,\y) {$s_{\r,t-1}$};
    \node[st] (b\r5) at (8.0,\y) {$s_{\r,t}$};
    \node[st] (b\r6) at (9.6,\y) {$s_{\r,t+1}$};
    \foreach \p/\q in {0/1,1/2,2/3,3/4,4/5}{\draw[tr] (b\r\p) -- (b\r\q);}
    \draw[tr, line width=1.1pt] (b\r5) -- (b\r6);
    \node[anchor=east] at (-0.75,\y) {agent $\r$};
  }
  \node at (4.85,-5.55) {(b) At time $t+1$};
\end{tikzpicture}
\caption{The $m$-step window moving from time $t$ to time $t+1$. Each row is the trajectory of one agent, and the arrows are transitions under its dynamics. (a)~At time $t$, the window covers times $t-m$ to $t-1$, and earlier visits form the pre-window part. (b)~After the transition to time $t+1$, the window shifts by one step, and the visits at time $t-m$ join the pre-window part. Pre-window curvature $c_m$ bounds how much pre-window visits, of any agent, reduce the marginal gain of a current visit.}
\label{fig:prewindow}
\end{figure}

Figure~\ref{fig:prewindow} illustrates the $m$-step window and the pre-window part on the trajectories of two agents.
\subsection{Properties of the Pre-Window Curvature}\label{app:cm_properties}
\begin{proposition}\label{prop:cm_properties}
For every $m\in\{0,\ldots,H\}$:
\begin{enumerate}[label=\roman*.]
 \item $0\leq c_m\leq c\leq1$;
 \item $c_{m+1}\leq c_m$ whenever $m<H$;
 \item $c_H=0$;
 \item if $F$ is additive across time, then $c_m=0$ for every $m$.
\end{enumerate}
\end{proposition}
\begin{proof}
\begin{enumerate}[label=\roman*.]
  \item We first show $c_m\geq0$ and $c\leq1$, and then show $c_m\leq c$. Definition~\ref{def:windowed_curvature} restricts the minimum to $x_t$ and $Z$ with $\Delta(x_t\mid Z)>0$, so every ratio is well defined. Since $Z\subseteq B\cup Z$, submodularity gives $\Delta(x_t\mid B\cup Z)\leq\Delta(x_t\mid Z)$, so every ratio is at most $1$ and $c_m\geq0$. In Definition~\ref{def:tot_curvature}, monotonicity makes the numerator of every ratio nonnegative, so every ratio is at least $0$ and $c\leq1$.

  To show $c_m\leq c$, we prove that every ratio in~\eqref{eq:window_curvature} is at least $1-c$. Fix $t$, $x_t\in\mathcal V_t$, $B\subseteq\mathcal V_{0:t-m-1}$, and $Z\subseteq\mathcal V$ with $\Delta(x_t\mid Z)>0$. The element $x_t$ does not lie in $Z$, since otherwise $\Delta(x_t\mid Z)=0$. It does not lie in $B$ either, since every element of $B$ has a time index $l<t-m\leq t$. Moreover, submodularity with $\emptyset\subseteq Z$ and $F(\emptyset)=0$ gives
  \begin{equation*}
   F(\{x_t\})=\Delta(x_t\mid\emptyset)\geq\Delta(x_t\mid Z)>0.
  \end{equation*}
  Hence $x_t\in\mathcal V\setminus(B\cup Z)$ and $F(\{x_t\})>0$. The ratio $\Delta(x_t\mid B\cup Z)/F(\{x_t\})$ is therefore one of the ratios minimized in Definition~\ref{def:tot_curvature}, so it is at least $1-c$. Since $1-c\geq0$ and $F(\{x_t\})\geq\Delta(x_t\mid Z)$, we obtain
  \begin{equation*}
   \frac{\Delta(x_t\mid B\cup Z)}{\Delta(x_t\mid Z)}
   \geq(1-c)\,\frac{F(\{x_t\})}{\Delta(x_t\mid Z)}
   \geq1-c.
  \end{equation*}
  The minimum in~\eqref{eq:window_curvature} is therefore at least $1-c$, so $c_m\leq c$.
  \item Since $\mathcal V_{0:t-m-2}\subseteq\mathcal V_{0:t-m-1}$, replacing $m$ by $m+1$ shrinks the set of admissible $B$ while the rest of the minimization is unchanged. The minimum over a smaller set can only increase, so $c_{m+1}\leq c_m$.
  \item For $m=H$ and any $t\in\{0,\ldots,H\}$, we have $t-H\leq0$, so $\mathcal V_{0:t-H-1}=\emptyset$ and $B=\emptyset$. Every ratio then equals $1$, and $c_H=0$.
  \item Suppose $F(X)=\sum_{t=0}^{H}f_t(X\cap\mathcal V_t)$. Adding $x_t\in\mathcal V_t$ changes only the term $f_t$, which depends on $X\cap\mathcal V_t$. Since $B\subseteq\mathcal V_{0:t-m-1}$ is disjoint from $\mathcal V_t$, we have $(B\cup Z)\cap\mathcal V_t=Z\cap\mathcal V_t$. Hence $\Delta(x_t\mid B\cup Z)=\Delta(x_t\mid Z)$, every ratio equals one, and $c_m=0$.
\end{enumerate}
\end{proof}

\section{Traffic-Monitoring Reward Properties}\label{app:traffic}
Proposition~\ref{prop:traffic} gives sufficient conditions on $\alpha_\ell$ under which the reward~\eqref{eq:traffic_reward} of Section~\ref{subsec:traffic_example} is monotone submodular and has $c_m<1$. Corollary~\ref{cor:traffic_submodular} shows that the exponential profile $\alpha_\ell=1-e^{-\beta\ell}$ satisfies these conditions, so the traffic-monitoring reward is monotone submodular. Proposition~\ref{prop:traffic_curvature} then computes $c_m$ in closed form for this profile.

\subsection{Discretely Concave Recovery Profiles}\label{subsec:app_concave}

We call the sequence $\boldsymbol\alpha=(\alpha_0,\ldots,\alpha_H)$ the recovery profile. In this subsection, $\boldsymbol\alpha$ is general rather than $\alpha_\ell=1-e^{-\beta\ell}$. We call $\boldsymbol\alpha$ discretely concave if
\begin{equation}\label{app_eq:discrete_concavity}
\alpha_{\ell+1}-\alpha_\ell\leq\alpha_\ell-\alpha_{\ell-1},\qquad 1\leq \ell<H.
\end{equation}
Each passing time step therefore recovers no more than the step before it. The concavity is strict when every inequality is strict.

\begin{proposition}\label{prop:traffic}
Suppose $\alpha_0=0$, $\alpha_\ell\in[0,1]$ for $1\leq\ell\leq H$, and $\boldsymbol\alpha$ is nondecreasing and discretely concave. Then the reward in~\eqref{eq:traffic_reward} is monotone submodular. If $\alpha_1>0$ and the concavity is strict, then $c_m<1$ for every $m<H$.
\end{proposition}

\begin{proof}
Fix $B\subseteq\mathcal V$ and consider adding $(t,s)$. The reward~\eqref{eq:traffic_reward} is a sum over roads, and each term depends on $B$ only through the monitoring times $T_e(B)$. Adding $(t,s)$ leaves $T_e(B)$ unchanged for $e\notin\mathcal C(s)$, and replaces it by $T_e(B)\cup\{t\}$ otherwise. Hence
\begin{equation*}
 \Delta\big((t,s)\mid B\big)
 =\sum_{e\in\mathcal C(s)}\Delta_e(t\mid B),
\end{equation*}
where $\Delta_e(t\mid B)$ is road $e$'s contribution. This contribution is zero if $t\in T_e(B)$. Otherwise, let $p$ and $q$ be the nearest monitoring times in $T_e(B)$ before and after $t$, when they exist. If $p$ exists, let $\ell_-=t-p$; if $q$ exists, let $\ell_+=q-t$. The four cases are
\begin{equation}\label{eq:app_traffic_marginal}
 \Delta_e(t\mid B)=w_e\begin{cases}
 1, & p,q\text{ do not exist},\\
 \alpha_{\ell_-}, & \text{only }p\text{ exists},\\
 \alpha_{\ell_+}, & \text{only }q\text{ exists},\\
 \alpha_{\ell_-}+\alpha_{\ell_+}-\alpha_{\ell_-+\ell_+}, & p,q\text{ both exist}.
 \end{cases}
\end{equation}
When both $p$ and $q$ exist, adding $(t,s)$ replaces the gap $\ell_-+\ell_+$ with the gaps $\ell_-$ and $\ell_+$.

Since the increments of $\boldsymbol\alpha$ are nonincreasing by~\eqref{app_eq:discrete_concavity}, $\alpha_{\ell_-+\ell_+}-\alpha_{\ell_+}\leq\alpha_{\ell_-}-\alpha_0=\alpha_{\ell_-}$. Moreover, $\boldsymbol\alpha$ is nondecreasing, so $\alpha_{\ell_-+\ell_+}\geq\max\{\alpha_{\ell_-},\alpha_{\ell_+}\}$. These inequalities give
\begin{equation}\label{eq:app_traffic_gap_property}
 0\leq\alpha_{\ell_-}+\alpha_{\ell_+}-\alpha_{\ell_-+\ell_+}\leq\min\{\alpha_{\ell_-},\alpha_{\ell_+}\}.
\end{equation}
The lower bound shows that the contribution in the last case of~\eqref{eq:app_traffic_marginal} is nonnegative. The upper bound shows that it is no greater than the corresponding contribution when only $p$ or only $q$ exists.

The four cases in~\eqref{eq:app_traffic_marginal} differ only in which neighboring monitoring times exist. The convention $\alpha_\infty=1$ of Section~\ref{subsec:traffic_example} removes this distinction. Set $\ell_-=\infty$ when $p$ does not exist and $\ell_+=\infty$ when $q$ does not exist, so that $\alpha_{\ell_-+\ell_+}=1$ whenever a gap is infinite. Every case of~\eqref{eq:app_traffic_marginal} then reads
\begin{equation}\label{eq:app_traffic_h}
 \Delta_e(t\mid B)=w_e\,h(\ell_-,\ell_+),\qquad h(\ell_1,\ell_2)=\alpha_{\ell_1}+\alpha_{\ell_2}-\alpha_{\ell_1+\ell_2},
\end{equation}
since $h(\infty,\infty)=1$, $h(\ell_1,\infty)=\alpha_{\ell_1}$, and $h(\infty,\ell_2)=\alpha_{\ell_2}$. The contribution of road $e$ therefore depends on $B$ only through the two gaps.

The function $h$ is nondecreasing in each argument over its domain. For finite $\ell_1,\ell_2$ with $\ell_1+\ell_2+1\leq H$,
\begin{equation*}
 h(\ell_1+1,\ell_2)-h(\ell_1,\ell_2)=(\alpha_{\ell_1+1}-\alpha_{\ell_1})-(\alpha_{\ell_1+\ell_2+1}-\alpha_{\ell_1+\ell_2})\geq0
\end{equation*}
by discrete concavity. When $\ell_2=\infty$, $h(\ell_1+1,\ell_2)-h(\ell_1,\ell_2) \geq 0$ follows from $h(\ell_1,\infty)=\alpha_{\ell_1}$ and the monotonicity of $\boldsymbol\alpha$. Moreover, $h(\ell_1,\ell_2)\leq\alpha_{\ell_2}=h(\infty,\ell_2)$, since $\alpha_{\ell_1}\leq\alpha_{\ell_1+\ell_2}$. The argument for $\ell_2$ is symmetric.

\emph{Monotonicity.} By~\eqref{eq:app_traffic_gap_property}, $h\geq0$ for finite gaps, and $h(\ell_1,\infty)=\alpha_{\ell_1}\geq0$. Hence $\Delta((t,s)\mid B)\geq0$, and $F$ is monotone.

\emph{Submodularity.} Let $B\subseteq B'$ and $(t,s)\notin B'$, and fix $e\in\mathcal C(s)$. If $t\in T_e(B')$, then $\Delta_e(t\mid B')=0\leq\Delta_e(t\mid B)$. Otherwise $t\notin T_e(B)$ as well, since $T_e(B)\subseteq T_e(B')$. Let $(\ell_-,\ell_+)$ and $(\ell'_-,\ell'_+)$ be the gaps under $T_e(B)$ and $T_e(B')$. Additional monitoring times can only move the nearest monitoring time on each side closer to $t$, and can only turn an absent neighbor into a present one, so $\ell'_-\leq\ell_-$ and $\ell'_+\leq\ell_+$. By~\eqref{eq:app_traffic_h} and the monotonicity of $h$,
\begin{equation*}
 \Delta_e(t\mid B')=w_e\,h(\ell'_-,\ell'_+)\leq w_e\,h(\ell_-,\ell_+)=\Delta_e(t\mid B).
\end{equation*}
Summing over $e\in\mathcal C(s)$ proves that $F$ is submodular.

\emph{Curvature.} Suppose $\alpha_1>0$ and the concavity is strict. Since $\boldsymbol\alpha$ is nondecreasing, $\alpha_\ell\geq\alpha_1>0$ for every positive gap $\ell$. For positive finite gaps $\ell_1,\ell_2$ with $\ell_1+\ell_2\leq H$, strict discrete concavity gives $\alpha_{\ell_1+\ell_2}-\alpha_{\ell_2}<\alpha_{\ell_1}-\alpha_0=\alpha_{\ell_1}$. Hence the contribution when both $p$ and $q$ exist is also positive.

Consider a ratio in~\eqref{eq:window_curvature} with $\Delta(x_t\mid Z)>0$, where $x_t=(t,s)$. At least one road $e\in\mathcal C(s)$ has a positive contribution under $Z$. Elements of $B$ occur before time $t-m$, so they cannot add $t$ to the monitoring times of road $e$. By~\eqref{eq:app_traffic_marginal}, this road still has a positive contribution under $B\cup Z$. Thus, the numerator is positive whenever the denominator is positive. The number of admissible ratios is finite, so their minimum is positive. Therefore, $c_m<1$.
\end{proof}

\begin{samepage}
\emph{Remark.} Non-strict discrete concavity alone does not guarantee $c_m<1$, even when $\alpha_1>0$. Fix $m\leq H-2$ and suppose
\begin{equation*}
 \alpha_{\ell+1}-\alpha_\ell=\alpha_1>0,\qquad 0\leq \ell\leq m+1.
\end{equation*}
Set $t=m+1$, $Z=\{(t+1,s)\}$, and $B=\{(0,s)\}$, where $s$ monitors at least one road with $w_e>0$. Since $t-m=1$, $B\subseteq\mathcal V_0$ is admissible. The marginal gain under $Z$ is positive. Under $B\cup Z$, the contribution of each road in $\mathcal C(s)$ is
\begin{equation*}
 w_e\bigl(\alpha_{m+1}+\alpha_1-\alpha_{m+2}\bigr)=0.
\end{equation*}
Hence the ratio in~\eqref{eq:window_curvature} is zero, and $c_m=1$.
\end{samepage}

For the exponential profile of Section~\ref{subsec:traffic_example}, Proposition~\ref{prop:traffic} gives the following result.

\begin{corollary}\label{cor:traffic_submodular}
For every $\beta>0$, the traffic-monitoring reward~\eqref{eq:traffic_reward} is monotone submodular.
\end{corollary}
\begin{proof}
The exponential profile $\alpha_\ell=1-e^{-\beta\ell}$ satisfies $\alpha_0=0$, is nondecreasing, and takes values in $[0,1]$. Its increments
\begin{equation*}
 \alpha_{\ell+1}-\alpha_\ell=e^{-\beta\ell}\big(1-e^{-\beta}\big)
\end{equation*}
are strictly decreasing in $\ell$, so $\boldsymbol\alpha$ is strictly discretely concave. Proposition~\ref{prop:traffic} gives the claim.
\end{proof}

\subsection{Pre-Window Curvature of the Exponential Profile}\label{subsec:app_exp}
\begin{proposition}\label{prop:traffic_curvature}
For every $\beta>0$, the traffic-monitoring reward~\eqref{eq:traffic_reward} has pre-window curvature
\begin{equation}\label{eq:exponential_curvature}
c_m=e^{-\beta(m+1)}, \qquad 0\leq m<H.
\end{equation}
\end{proposition}
\begin{proof}
We first show that every ratio $\Delta(x_t\mid B\cup Z)/\Delta(x_t\mid Z)$ in~\eqref{eq:window_curvature} is at least $\alpha_{m+1}$, and then show that this bound is attained. Fix $x_t=(t,s)$ and a road $e\in\mathcal C(s)$. Let $\ell_-$ and $\ell_+$ be the gaps to the nearest monitoring times of $e$ in $Z$ before and after $t$, and let $\ell_B$ be the gap to the nearest monitoring time of $e$ in $B$. Set a gap to infinity when the corresponding monitoring time does not exist. Since $B$ contains only elements before time $t-m$, $\ell_B\geq m+1$. Thus, adding $B$ changes $\ell_-$ to $\min\{\ell_-,\ell_B\}$ and leaves $\ell_+$ unchanged. Figure~\ref{fig:app_gaps} illustrates these monitoring times.

\begin{figure}[t]
\centering
\begin{tikzpicture}[x=0.95cm,y=0.95cm,font=\footnotesize]
  \fill[black!8] (4.2,-0.16) rectangle (5.4,0.16);
  \node[font=\scriptsize,black!60] at (4.8,-0.58) {$m$-step window};
  \draw[->,gray!70] (0.2,0) -- (8.3,0);
  \node[font=\scriptsize,gray!70] at (8.3,-0.32) {time};
  \foreach \x in {0.8,2.6,4.2,5.4,7.4}{\draw[gray!70] (\x,-0.09) -- (\x,0.09);}
  \fill (0.8,0) circle (2.2pt);
  \draw[fill=white,thick] (2.525,-0.075) rectangle (2.675,0.075);
  \fill (7.4,0) circle (2.2pt);
  \draw[thick] (5.3,-0.10)--(5.5,0.10) (5.3,0.10)--(5.5,-0.10);
  \node[below=3pt] at (4.2,-0.02) {$t-m$};
  \node[below=3pt] at (5.4,0) {$t$};
  \draw[<->] (0.8,1.15) -- node[above,inner sep=1.5pt]{$\ell_-$} (5.4,1.15);
  \draw[<->] (2.6,0.62) -- node[above,inner sep=1.5pt]{$\ell_B$} (5.4,0.62);
  \draw[<->] (5.4,0.62) -- node[above,inner sep=1.5pt]{$\ell_+$} (7.4,0.62);
  \draw[gray!45,dashed] (0.8,0.12) -- (0.8,1.12);
  \draw[gray!45,dashed] (2.6,0.12) -- (2.6,0.59);
  \draw[gray!45,dashed] (5.4,0.20) -- (5.4,1.12);
  \draw[gray!45,dashed] (7.4,0.12) -- (7.4,0.59);
\end{tikzpicture}
\caption{Monitoring times of road $e$ around the added element $x_t=(t,s)$. Filled circles denote monitoring times from $Z$, the open square denotes the nearest monitoring time from $B$, and the cross denotes $x_t$. Elements of $B$ occur before time $t-m$, so adding $B$ changes the earlier gap from $\ell_-$ to $\min\{\ell_-,\ell_B\}$ and leaves the later gap $\ell_+$ unchanged.}
\label{fig:app_gaps}
\end{figure}
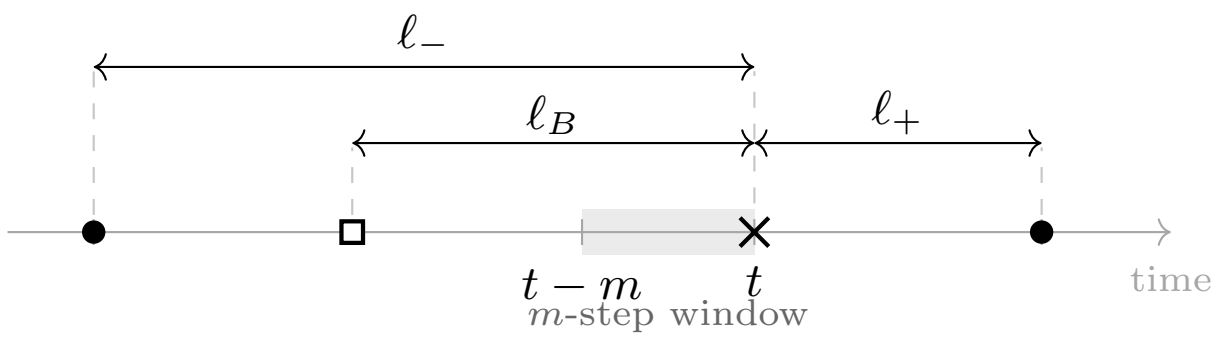

For finite gaps $\ell_-$ and $\ell_+$, the exponential profile satisfies
\begin{equation*}
 \alpha_{\ell_-}+\alpha_{\ell_+}-\alpha_{\ell_-+\ell_+}=(1-e^{-\beta\ell_-})(1-e^{-\beta\ell_+})=\alpha_{\ell_-}\alpha_{\ell_+}.
\end{equation*}
With the convention $\alpha_\infty=1$, the four cases of~\eqref{eq:app_traffic_marginal} for $t\notin T_e(Z)$ can therefore be written as
\begin{equation}\label{eq:app_exp_factorization}
 \Delta_e(t\mid Z)=w_e\alpha_{\ell_-}\alpha_{\ell_+}.
\end{equation}
Consider any admissible $(t,x_t,B,Z)$ in Definition~\ref{def:windowed_curvature}, so that $\Delta(x_t\mid Z)>0$. For each road with $\Delta_e(t\mid Z)>0$,~\eqref{eq:app_exp_factorization} gives
\begin{equation*}
 \frac{\Delta_e(t\mid B\cup Z)}{\Delta_e(t\mid Z)}=\frac{\alpha_{\min\{\ell_-,\ell_B\}}}{\alpha_{\ell_-}}
 =\begin{cases}
 1, & \ell_B\geq\ell_-,\\
 \alpha_{\ell_B}/\alpha_{\ell_-}, & \ell_B<\ell_-.
 \end{cases}
\end{equation*}
Both cases are at least $\alpha_{m+1}$: the first follows from $\alpha_{m+1}\leq1$, and the second from $\alpha_{\ell_-}\leq1$ and $\alpha_{\ell_B}\geq\alpha_{m+1}$.

To extend this bound to the complete marginal gain, let $E^+=\{e\in\mathcal C(s):\Delta_e(t\mid Z)>0\}$. A road with zero contribution under $Z$ also has zero contribution under $B\cup Z$. Therefore,
\begin{align}
 \frac{\Delta(x_t\mid B\cup Z)}{\Delta(x_t\mid Z)}
 &=\frac{\sum_{e\in E^+}\Delta_e(t\mid B\cup Z)}{\sum_{e\in E^+}\Delta_e(t\mid Z)}\notag\\
 &\geq\min_{e\in E^+}\frac{\Delta_e(t\mid B\cup Z)}{\Delta_e(t\mid Z)}\geq\alpha_{m+1}.
 \label{eq:app_exp_mediant}
\end{align}
Hence every admissible ratio is at least $\alpha_{m+1}$, and $c_m\leq1-\alpha_{m+1}$.

For $m<H$, set $t=m+1$, $Z=\emptyset$, and $B=\{(0,s)\}$, where $s$ monitors at least one road with $w_e>0$. Since $t-m=1$, the set $B\subseteq\mathcal V_0$ is admissible. Under $Z$, $x_t=(t,s)$ is the first monitoring of every road in $\mathcal C(s)$, whereas under $B$ it follows the previous monitoring by $m+1$ time steps. Thus
\begin{equation*}
 \frac{\Delta(x_t\mid B)}{\Delta(x_t\mid Z)}
 =\alpha_{m+1}.
\end{equation*}
The lower bound is attained, so
\begin{equation*}
 c_m=1-\alpha_{m+1}=e^{-\beta(m+1)}.
\end{equation*}
\end{proof}
At $m=H$, no visit can occur before the $H$-step window, and $c_H=0$.

\section{Analysis of the Windowed Objective}
\subsection{Set Form of the Pre-Window Curvature Inequality}
Definition~\ref{def:windowed_curvature} bounds the effect of visits prior to the window on the marginal gain of a single visit $x_t\in\mathcal V_t$ given a set $Z$. The proof of Lemma~\ref{lem:sandwich} needs the same bound for a set of visits at time $t$, so we extend it to any set $X_t\subseteq\mathcal V_t$.
\begin{lemma}\label{lem:curvature_set_extension}
For every $m, t \in \{0, \dots, H\}$, $X_t \subseteq \mathcal{V}_t$, $B \subseteq \mathcal V_{0:t-m-1}$, and $Z \subseteq \mathcal{V}$,
\begin{equation*}
\Delta\big(X_t \mid B \cup Z\big) \;\geq\; (1-c_m)\, \Delta\big(X_t \mid Z\big).
\end{equation*}
\end{lemma}
\begin{proof}
Elements of $X_t\cap Z$ add zero marginal gain on both sides, so it suffices to consider the elements of $X_t\setminus Z$, enumerated as $x_1,\ldots,x_q$. No element of $X_t$ lies in $B$, since $X_t\subseteq\mathcal V_t$ while $B$ contains only elements of earlier times. Let $Z_j=Z\cup\{x_1,\ldots,x_{j-1}\}$ collect the elements added before $x_j$. Both marginal gains then decompose into single-element gains whose $j$-th terms differ only in the presence of the set $B$:
\begin{align*}
 \Delta(X_t\mid B\cup Z)&=\sum_{j=1}^q\Delta(x_j\mid B\cup Z_j),\\
 \Delta(X_t\mid Z)&=\sum_{j=1}^q\Delta(x_j\mid Z_j).
\end{align*}
Each pair of terms is an instance of~\eqref{eq:window_curvature} with $x_t=x_j$, the same $B$, and $Z=Z_j$. Hence $\Delta(x_j\mid B\cup Z_j)\geq(1-c_m) \Delta(x_j\mid Z_j)$. When $\Delta(x_j\mid Z_j)=0$, submodularity and $Z_j\subseteq B\cup Z_j$ force $\Delta(x_j\mid B\cup Z_j)=0$, so the inequality holds trivially. Summing over $j$ proves the lemma.
\end{proof}

\subsection{Proofs of Lemma~\ref{lem:sandwich} and Theorem~\ref{thm:surrogate_guarantee}}\label{app:windowed_sandwich}
We first prove Lemma~\ref{lem:sandwich}, restated below.
\WindowedSandwich*
\begin{proof}
Fix a realized joint trajectory $\boldsymbol{\tau}$. Telescoping its
marginal gains in chronological order gives
\begin{equation*}
F(\boldsymbol{\tau})
=
\sum_{t=0}^{H}
\Delta(\boldsymbol{\tau}_t\mid\boldsymbol{\tau}_{0:t-1}).
\end{equation*}

For every $t$, the full past decomposes as
$\boldsymbol{\tau}_{0:t-1}=\boldsymbol{\tau}_{0:t-m-1}\cup\boldsymbol{\tau}_{t-m:t-1}$, where $\boldsymbol{\tau}_{0:t-m-1}$ is empty when $t\leq m$. For the lower bound, we apply Lemma~\ref{lem:curvature_set_extension} with $X_t=\boldsymbol{\tau}_t\subseteq\mathcal V_t$, $B=\boldsymbol{\tau}_{0:t-m-1}\subseteq\mathcal V_{0:t-m-1}$, and $Z=\boldsymbol{\tau}_{t-m:t-1}$. Since $B\cup Z=\boldsymbol{\tau}_{0:t-1}$, this gives
\begin{equation*}
\Delta(\boldsymbol{\tau}_t\mid\boldsymbol{\tau}_{0:t-1})
\geq
(1-c_m)\,\Delta(\boldsymbol{\tau}_t\mid\boldsymbol{\tau}_{t-m:t-1}).
\end{equation*}
For the upper bound, submodularity and $\boldsymbol{\tau}_{t-m:t-1}\subseteq\boldsymbol{\tau}_{0:t-1}$ give
\begin{equation*}
\Delta(\boldsymbol{\tau}_t\mid\boldsymbol{\tau}_{0:t-1})
\leq
\Delta(\boldsymbol{\tau}_t\mid\boldsymbol{\tau}_{t-m:t-1}).
\end{equation*}
Summing both bounds over $t$ and taking expectations under $\boldsymbol\pi$ proves the lemma. The left-hand sides give $J(\boldsymbol\pi)$ by the telescoping identity above, and the terms $\Delta(\boldsymbol\tau_t\mid\boldsymbol\tau_{t-m:t-1})=r_t(\tilde{\boldsymbol s}_t)$ give $\Phi_m(\boldsymbol\pi)$.
\end{proof}

We then prove Theorem~\ref{thm:surrogate_guarantee}, restated below.
\SurrogateGuarantee*
\begin{proof}
Applying Lemma~\ref{lem:sandwich} to $\boldsymbol{\pi}^*$ and $\boldsymbol{\pi}'$, and using the optimality of $\boldsymbol{\pi}'$ for $\Phi_m$, gives
\begin{equation*}
\begin{aligned}
(1-c_m)J^* &\leq (1-c_m)\Phi_m(\boldsymbol{\pi}^*)\\
  &\leq (1-c_m)\Phi_m(\boldsymbol{\pi}') \leq J(\boldsymbol{\pi}').
\end{aligned}\qedhere
\end{equation*}
\end{proof}

The guarantee does not require an exact maximizer of $\Phi_m$. Suppose that a policy $\boldsymbol\pi$ satisfies $\Phi_m(\boldsymbol\pi)\geq\Phi_m(\boldsymbol\pi')-\eta$ for some $\eta\geq0$. Combining the first two steps above with the lower bound of Lemma~\ref{lem:sandwich} for $\boldsymbol\pi$ gives
\begin{equation*}
 (1-c_m)J^*\leq(1-c_m)\Phi_m(\boldsymbol\pi')\leq(1-c_m)\big(\Phi_m(\boldsymbol\pi)+\eta\big)\leq J(\boldsymbol\pi)+(1-c_m)\eta,
\end{equation*}
so $J(\boldsymbol\pi)\geq(1-c_m)(J^*-\eta)$.

\subsection{Backward Induction on the Augmented State Space}\label{app:augmented_dp}
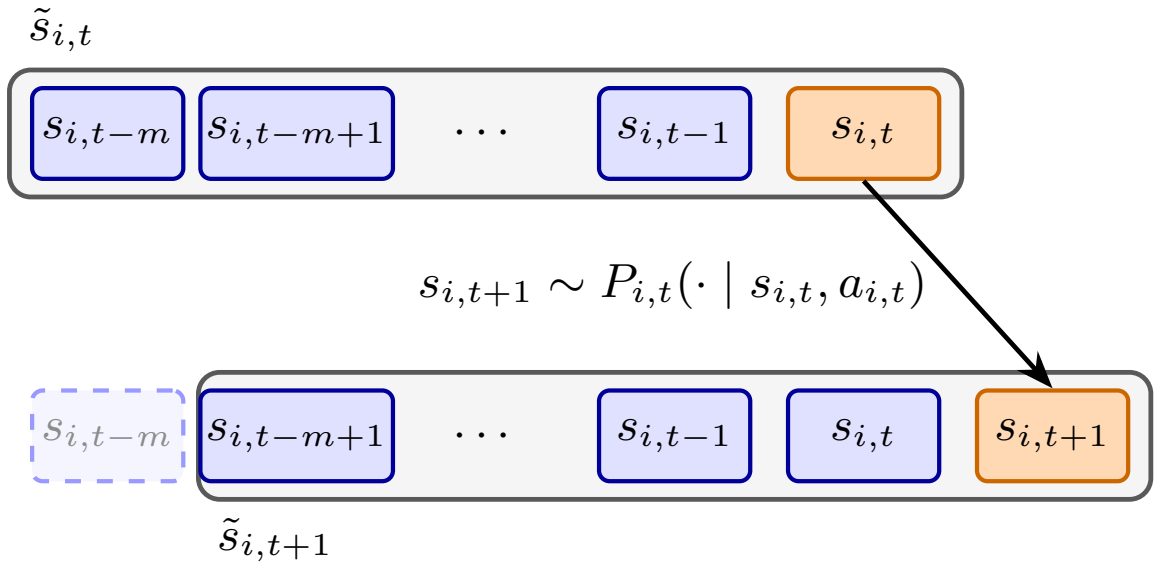
\begin{figure}[t]
\centering
\begin{tikzpicture}[
  font=\footnotesize,
  st/.style={draw, rounded corners=2pt, minimum width=1cm, minimum height=0.6cm,
             inner sep=1pt, line width=0.8pt},
  hist/.style={st, draw=blue!60!black, fill=blue!12},
  cur/.style={st, draw=orange!80!black, fill=orange!30},
  old/.style={st, dashed, draw=blue!40, fill=blue!4, text=black!45},
  grp/.style={draw=black!65, fill=black!4, rounded corners=4pt, line width=0.9pt},
]
  \draw[grp] (-0.65,-0.42) rectangle (5.65,0.42);
  \draw[grp] (0.6,-2.42) rectangle (6.9,-1.58);
  \node[hist] at (0,0)    {$s_{i,t-m}$};
  \node[hist] at (1.25,0) {$s_{i,t-m+1}$};
  \node       at (2.5,0)  {$\cdots$};
  \node[hist] at (3.75,0) {$s_{i,t-1}$};
  \node[cur] (cur) at (5,0) {$s_{i,t}$};
  \node[old]  at (0,-2)    {$s_{i,t-m}$};
  \node[hist] at (1.25,-2) {$s_{i,t-m+1}$};
  \node       at (2.5,-2)  {$\cdots$};
  \node[hist] at (3.75,-2) {$s_{i,t-1}$};
  \node[hist] at (5,-2)    {$s_{i,t}$};
  \node[cur] (next) at (6.25,-2) {$s_{i,t+1}$};
  \node[anchor=south west] at (-0.65,0.42) {$\tilde s_{i,t}$};
  \node[anchor=north west] at (0.6,-2.42)  {$\tilde s_{i,t+1}$};
  \draw[-{Stealth}, line width=0.9pt] (cur.south) -- (next.north)
    node[midway, left, xshift=-2pt] {$s_{i,t+1}\sim P_{i,t}(\cdot\mid s_{i,t},a_{i,t})$};
\end{tikzpicture}
\caption{History-augmented state of agent $i$ with memory length $m$, for $t\geq m$. The augmented state $\tilde s_{i,t}$ contains the history $h_{i,t}$ (blue) and the current state $s_{i,t}$ (orange). The next state $s_{i,t+1}$ is drawn from $P_{i,t}$, and the history appends $s_{i,t}$ and drops the oldest state $s_{i,t-m}$ (dashed).}
\label{fig:augmented_state}
\end{figure}

Algorithm~\ref{alg:augmented_dp} solves a finite-horizon MDP on the history-augmented states by backward induction~\citep{puterman1994}. The expectation is over the next state, while the history updates deterministically through the map $\varphi:\mathcal H^m_t\times\mathcal S\to\mathcal H^m_{t+1}$, which appends a state to a history and drops the first entry if the result has more than $m$ entries. Figure~\ref{fig:augmented_state} illustrates this update for one agent. Section~\ref{subsec:surrogate} applies the algorithm to the joint augmented states, with $\mathcal S$ and $\mathcal A$ replaced by $\mathbf S$ and $\mathbf A$. Section~\ref{subsec:sequential} applies it once per agent, with the estimated rewards $\widehat r_{i,t}^{\bar{\boldsymbol\pi}_{[i-1]}}$.
\begin{algorithm}[t]
\caption{Backward Induction on the Augmented State Space}
\label{alg:augmented_dp}
\begin{algorithmic}[1]
\REQUIRE spaces $\mathcal S,\mathcal A$; memory length $m$;
  dynamics $P=(P_0,\ldots,P_{H-1})$; rewards $r_t$
\STATE $V_H(\tilde s)\gets r_H(\tilde s)$ \quad $\forall\tilde s\in\tilde{\mathcal S}_H$
\FOR{$t=H-1,\ldots,0$}
  \FORALL{$\tilde s=(s,h)\in\tilde{\mathcal S}_t$ and $a\in\mathcal A$}
    \STATE $Q_t(\tilde s,a)\gets r_t(\tilde s)
      +\mathbb E_{s'\sim P_t(\cdot\mid s,a)}
      \big[V_{t+1}(s',\varphi(h,s))\big]$
  \ENDFOR
  \STATE $V_t(\tilde s)\gets\max_a Q_t(\tilde s,a)$ \quad $\forall\tilde s\in\tilde{\mathcal S}_t$
  \STATE $\pi_t(\tilde s)\in\arg\max_a Q_t(\tilde s,a)$ \quad $\forall\tilde s\in\tilde{\mathcal S}_t$
\ENDFOR
\RETURN $\pi=(\pi_0,\ldots,\pi_{H-1})$
\end{algorithmic}
\end{algorithm}

Backward induction on the joint augmented MDP computes a $Q$-value for each of the at most $|\mathcal S|^{N(m+1)}|\mathcal A|^N$ pairs in $\tilde{\mathbf S}_t\times\mathbf A$ at every stage. Each $Q$-value sums over at most $|\mathcal S|^N$ next joint physical states, since the histories update deterministically through $\varphi$. Over $H$ stages, the total computation time is $O(H|\mathcal S|^{N(m+2)}|\mathcal A|^N)$.

\subsection{Non-Monotonicity of the Windowed Stage Reward}\label{app:windowed_nonmonotonicity}
\begin{figure}[t]
\centering
\begin{tikzpicture}[x=1.3cm,y=0.85cm,font=\small,
  vis/.style={circle,draw,minimum size=6mm,inner sep=0pt},
  mon/.style={vis,fill=black!18},
  key/.style={mon,line width=1.4pt}]
  \begin{scope}
    \fill[black!7] (0.55,-0.5) rectangle (1.45,1.5);
    \node[font=\scriptsize,black!60] at (1,1.8) {window at $t=2$};
    \draw[->,gray!70] (-0.4,-0.8) -- (2.5,-0.8);
    \foreach \t in {0,1,2}{
      \draw[gray!70] (\t,-0.72) -- (\t,-0.88);
      \node[below,inner sep=2pt] at (\t,-0.88) {$\t$};
    }
    \node[font=\scriptsize,gray!70,anchor=west] at (2.5,-0.8) {time};
    \node[anchor=east] at (-0.45,1) {agent 1};
    \node[mon] at (0,1) {$a$};
    \node[vis] at (1,1) {$b$};
    \node[mon] at (2,1) {$a$};
    \node[anchor=east] at (-0.45,-1.75) {$r_t(\tilde{\boldsymbol s}_t)$};
    \node at (0,-1.75) {$1$};
    \node at (1,-1.75) {$0$};
    \node at (2,-1.75) {$1$};
    \node at (1,-2.5) {$\sum_t r_t(\tilde{\boldsymbol s}_t)=2$,\quad $F(\tau_1)=23/16$};
    \node at (1,-3.2) {(a) Agent 1 alone};
  \end{scope}
  \begin{scope}[xshift=6.5cm]
    \fill[black!7] (0.55,-0.5) rectangle (1.45,1.5);
    \node[font=\scriptsize,black!60] at (1,1.8) {window at $t=2$};
    \draw[->,gray!70] (-0.4,-0.8) -- (2.5,-0.8);
    \foreach \t in {0,1,2}{
      \draw[gray!70] (\t,-0.72) -- (\t,-0.88);
      \node[below,inner sep=2pt] at (\t,-0.88) {$\t$};
    }
    \node[font=\scriptsize,gray!70,anchor=west] at (2.5,-0.8) {time};
    \node[anchor=east] at (-0.45,1) {agent 1};
    \node[anchor=east] at (-0.45,0) {agent 2};
    \node[mon] at (0,1) {$a$};
    \node[vis] at (1,1) {$b$};
    \node[mon] at (2,1) {$a$};
    \node[vis] at (0,0) {$b$};
    \node[key] at (1,0) {$a$};
    \node[vis] at (2,0) {$b$};
    \node[anchor=east] at (-0.45,-1.75) {$r_t(\tilde{\boldsymbol s}_t)$};
    \node at (0,-1.75) {$1$};
    \node at (1,-1.75) {$1/4$};
    \node at (2,-1.75) {$1/4$};
    \node at (1,-2.5) {$\sum_t r_t(\tilde{\boldsymbol s}_t)=3/2$,\quad $F(\tau_1\cup\tau_2)=3/2$};
    \node at (1,-3.2) {(b) Both agents};
  \end{scope}
\end{tikzpicture}
\caption{Adding an agent lowers the windowed objective in the example of Appendix~\ref{app:windowed_nonmonotonicity}. Shaded nodes are visits to state $a$, which monitors road $e$. The gray region denotes the window at time $2$, and the values below the axis are the stage rewards, which sum to the windowed objective. (a)~With agent~1 alone, the window at time $2$ contains no visit to $a$. (b)~With both agents, the window at time $2$ contains the visit $(1,a)$ of agent~2. This visit lowers the stage reward at time $2$, although the true reward $F$ increases.}
\label{fig:app_nonmonotone}
\end{figure}
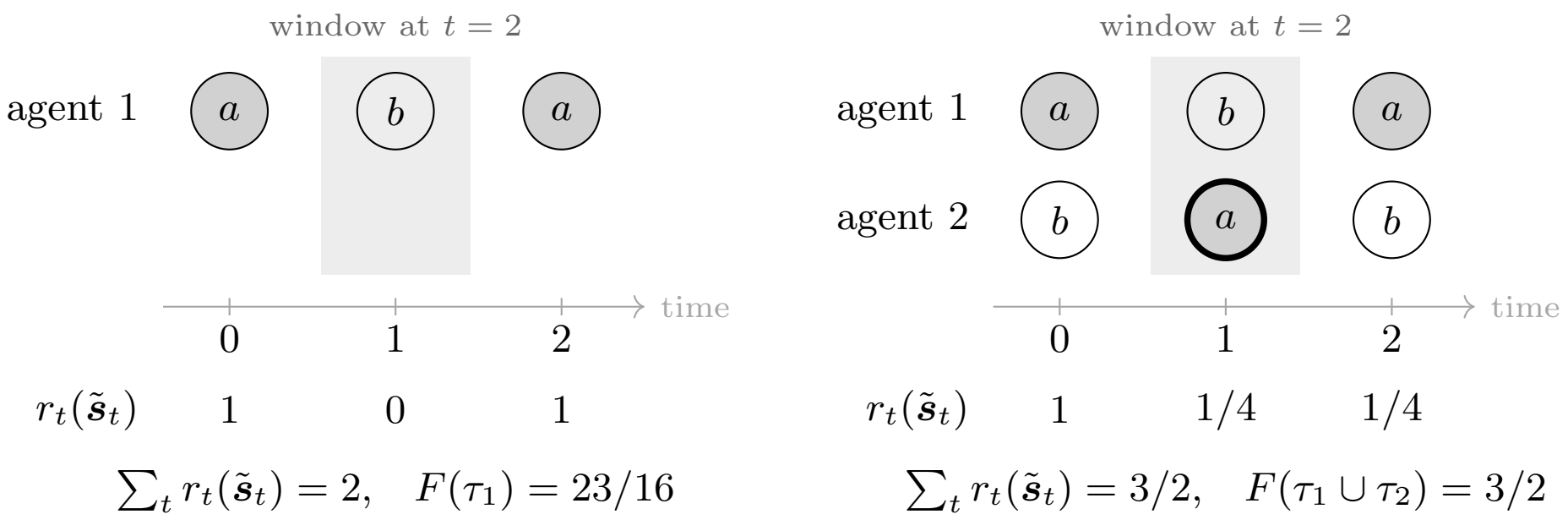

Adding an agent can decrease the windowed objective even when $F$ is monotone submodular and $c_m<1$. Consider a reward of the form~\eqref{eq:traffic_reward} on two states $a$ and $b$ with a single road $e$ of maximum importance $w_e=1$, where $\mathcal C(a)=\{e\}$ and $\mathcal C(b)=\emptyset$. Set $H=2$, $m=1$, and $\beta=\ln(4/3)$, so that $\alpha_1=1/4$ and $\alpha_2=7/16$. By Corollary~\ref{cor:traffic_submodular}, the reward is monotone submodular, and~\eqref{eq:exponential_curvature} gives $c_1=9/16$. Each agent can move between $a$ and $b$ deterministically, and two agents with fixed initial states follow the trajectories
\begin{equation*}
 \tau_1=\{(0,a),(1,b),(2,a)\},\qquad \tau_2=\{(0,b),(1,a),(2,b)\},
\end{equation*}
shown in Figure~\ref{fig:app_nonmonotone}. Since the dynamics and initial states are deterministic, the windowed objective of the policies that generate these trajectories equals the sum of the stage rewards $r_t(\tilde{\boldsymbol s}_t)$ along them.

With agent~$1$ alone, its visit at time $0$ lies outside the window at time $2$. The stage rewards are then $r_0(\tilde{\boldsymbol s}_0)=1$, $r_1(\tilde{\boldsymbol s}_1)=0$, and $r_2(\tilde{\boldsymbol s}_2)=1$, which sum to $2$. With both agents, road $e$ is monitored at every time. The stage rewards are then $r_0(\tilde{\boldsymbol s}_0)=1$, $r_1(\tilde{\boldsymbol s}_1)=\alpha_1$, and $r_2(\tilde{\boldsymbol s}_2)=\alpha_1$, which sum to $3/2$. Adding agent~$2$ thus lowers the stage reward at time $2$ from $1$ to $1/4$, because its visit $(1,a)$ enters the window at time $2$. The true reward instead increases, from $F(\tau_1)=1+\alpha_2=23/16$ to $F(\tau_1\cup\tau_2)=3/2$. Hence, the stage rewards of $\Phi_m$ are not monotone in the set of agents, and the condition required by the guarantee of~\citet{chen2026} fails.

\section{Analysis of Sequential Policy Optimization}
\label{app:sequential}

We prove Theorem~\ref{thm:sequential_guarantee}, restated below.
\SequentialGuarantee*

Throughout this section, we use the following notation. For a joint policy $\boldsymbol\pi$ and a local policy $\pi_i$ of agent $i$, let $d_{i,t}^{\boldsymbol\pi},d_{i,t}^{\pi_i}\in\mathcal D(\tilde{\mathcal S}_t)$ denote the distributions of the augmented state of agent $i$ at time $t$,
\begin{equation*}
 d_{i,t}^{\boldsymbol\pi}(\tilde s)=\Pr\nolimits_{\boldsymbol\tau\sim\boldsymbol\pi}\big(\tilde s_{i,t}=\tilde s\big),
 \qquad
 d_{i,t}^{\pi_i}(\tilde s)=\Pr\nolimits_{\tau_i\sim\pi_i}\big(\tilde s_{i,t}=\tilde s\big),
\end{equation*}
for every $\tilde s\in\tilde{\mathcal S}_t$. With this notation, the local objective~\eqref{eq:G_objective} takes the equivalent form
\begin{equation}\label{eq:G_occupancy}
 G_i^{\bar{\boldsymbol\pi}_{[i-1]}}(\pi_i)
 =\sum_{t=0}^H
 \mathbb E_{\tilde s_{i,t}\sim d_{i,t}^{\pi_i}}
 \left[r_{i,t}^{\bar{\boldsymbol\pi}_{[i-1]}}(\tilde s_{i,t})\right].
\end{equation}

\subsection{Proof of Theorem~\ref{thm:sequential_guarantee}}
\label{app:thm2}

\begin{proof}
The argument has three steps. Step~1 bounds the optimality gap by the sum of the maximum local objectives of the agents. Step~2 bounds each maximum by the attained local objective, up to the reward-estimation error. Step~3 bounds the attained local objectives by the return $J(\bar{\boldsymbol\pi})$ through $c_m$.

The policy $\bar{\boldsymbol\pi}$ returned by Algorithm~\ref{alg:GWPO} depends on the sampled predecessor trajectories. Throughout the proof, we treat $\bar{\boldsymbol\pi}$ as fixed. Since each $\bar\pi_i$ is deterministic, an expectation over $\boldsymbol\tau\sim\bar{\boldsymbol\pi}$ is then taken over the initial states and the transitions only.

\emph{Step 1 (bounding the optimality gap by the maximum local objectives).} For $\boldsymbol{\tau}^* \sim \boldsymbol{\pi}^*$ and $\boldsymbol{\tau} \sim \bar{\boldsymbol{\pi}}$ sampled independently, with $\boldsymbol{\pi}^*$ an optimal global history-dependent policy of \eqref{eq:objective} and $\bar{\boldsymbol{\pi}}\in\boldsymbol\Pi^{\mathrm{loc}}_m$ the policy returned by Algorithm~\ref{alg:GWPO}, we have
\begin{align*}
J^* = \mathbb{E}_{\boldsymbol{\tau}^* \sim \boldsymbol{\pi}^*}[F(\boldsymbol{\tau}^*)] \leq \mathbb{E}_{\boldsymbol{\tau}^* \sim \boldsymbol{\pi}^*, \boldsymbol{\tau} \sim \bar{\boldsymbol{\pi}}}[F(\boldsymbol{\tau}^* \cup \boldsymbol{\tau})] &= \mathbb{E}_{\boldsymbol{\tau}^* \sim \boldsymbol{\pi}^*, \boldsymbol{\tau} \sim \bar{\boldsymbol{\pi}}}[F(\boldsymbol{\tau}) + \Delta(\boldsymbol{\tau}^* \mid \boldsymbol{\tau})]\\
&= \mathbb{E}_{\boldsymbol{\tau} \sim \bar{\boldsymbol{\pi}}}[F(\boldsymbol{\tau})] + \mathbb{E}_{\boldsymbol{\tau}^* \sim \boldsymbol{\pi}^*, \boldsymbol{\tau} \sim \bar{\boldsymbol{\pi}}}[\Delta(\boldsymbol{\tau}^* \mid \boldsymbol{\tau})]\\
&= J(\bar{\boldsymbol{\pi}}) + \mathbb{E}_{\boldsymbol{\tau}^* \sim \boldsymbol{\pi}^*, \boldsymbol{\tau} \sim \bar{\boldsymbol{\pi}}}[\Delta(\boldsymbol{\tau}^* \mid \boldsymbol{\tau})],
\end{align*}
where the inequality holds by monotonicity, and the equalities use the definition of the marginal gain, linearity of expectation, and the definition of the return \eqref{eq:objective}. For every fixed realization of $\boldsymbol{\tau}$ and $\boldsymbol{\tau}^*$, the pairs $\tau^*_{i,t}$ of $\boldsymbol{\tau}^*$ are added to $\boldsymbol{\tau}$ one at a time in lexicographic order of $(i,t)$, meaning $\boldsymbol{\tau}^*_{[i-1]} \cup \tau^*_{i, 0:t-1}$ is the set of elements added before $\tau^*_{i,t}$. Telescoping these marginal gains gives
\begin{equation*}
\Delta(\boldsymbol{\tau}^* \mid \boldsymbol{\tau}) = \sum_{i \in [N]} \sum_{t=0}^{H} \Delta\big( \tau^*_{i,t} \mid \boldsymbol{\tau} \cup \boldsymbol{\tau}^*_{[i-1]} \cup \tau^*_{i, 0:t-1} \big),
\end{equation*}
where a term whose pair already lies in its given set, $\tau^*_{i,t} \subseteq \boldsymbol{\tau} \cup \boldsymbol{\tau}^*_{[i-1]} \cup \tau^*_{i, 0:t-1}$, contributes $0$, since $\Delta(x \mid B) = 0$ whenever $x \in B$. The set $\boldsymbol{\tau} \cup \boldsymbol{\tau}^*_{[i-1]} \cup \tau^*_{i, 0:t-1}$ given in each term contains $\tau^*_{i, t-m:t-1} \cup \boldsymbol{\tau}_{[i-1]}$, because $\boldsymbol{\tau}_{[i-1]} \subseteq \boldsymbol{\tau}$ and $\tau^*_{i, t-m:t-1} \subseteq \tau^*_{i, 0:t-1}$. By submodularity, the marginal gain of $\tau^*_{i,t}$ does not decrease when the given set shrinks to $\tau^*_{i,t-m:t-1}\cup\boldsymbol\tau_{[i-1]}$, so
\begin{equation*}
\Delta(\boldsymbol{\tau}^* \mid \boldsymbol{\tau}) \leq \sum_{i \in [N]} \sum_{t=0}^{H} \Delta\big( \tau^*_{i,t} \,\big|\, \tau^*_{i, t-m:t-1} \cup \boldsymbol{\tau}_{[i-1]} \big).
\end{equation*}
It remains to take the expectation of this bound over the two runs. Along the optimal run, the augmented state $\tilde s^*_{i,t}=(s^*_{i,t},h^*_{i,t})$ determines $\tau^*_{i,t}$ and $\tau^*_{i,t-m:t-1}=\nu_t(h^*_{i,t})$, so each term depends on agent $i$'s trajectory only through $\tilde s^*_{i,t}$. Since $\tilde s^*_{i,t} \sim d^{\boldsymbol{\pi}^*}_{i,t}$ is independent of the predecessors' set $\boldsymbol{\tau}_{[i-1]}$, the expectation of each term splits into an outer expectation over $\tilde s^*_{i,t}$ and an inner expectation over $\boldsymbol{\tau}_{[i-1]}$:
\begin{align*}
\mathbb{E}_{\boldsymbol{\tau}^* \sim \boldsymbol{\pi}^*, \boldsymbol{\tau} \sim \bar{\boldsymbol{\pi}}}\Big[\Delta\big( \tau^*_{i,t} \,\big|\, \tau^*_{i, t-m:t-1} \cup \boldsymbol{\tau}_{[i-1]} \big)\Big]
&= \mathbb{E}_{\tilde s^*_{i,t} \sim d^{\boldsymbol{\pi}^*}_{i,t}}\Big[\, \mathbb{E}_{\boldsymbol\tau_{[i-1]}\sim\bar{\boldsymbol\pi}_{[i-1]}}\Big[ \Delta\big((t, s^*_{i,t}) \mid \nu_t(h^*_{i,t}) \cup \boldsymbol{\tau}_{[i-1]}\big) \Big] \Big]\\
&= \mathbb{E}_{\tilde s^*_{i,t} \sim d^{\boldsymbol{\pi}^*}_{i,t}}\big[ r_{i,t}^{\bar{\boldsymbol\pi}_{[i-1]}}(\tilde s^*_{i,t}) \big],
\end{align*}
where the second equality applies the definition \eqref{eq:local_reward} of the reward. The occupancy $d^{\boldsymbol{\pi}^*}_{i,t}$ is induced under the global history-dependent policy $\boldsymbol{\pi}^*$, whereas the algorithm optimizes over local Markov policies. The following lemma closes this gap.

\begin{restatable}{lemma}{MarginalRealizability}\label{lem:realizability}
Fix an agent $i$ and any $\boldsymbol\pi \in \boldsymbol\Pi_H$. There is a policy $\pi_i''\in\Pi^{\mathrm{loc}}_m$ with $d_{i,t}^{\pi_i''}=d_{i,t}^{\boldsymbol\pi}$ for every $t$.
\end{restatable}

Applying Lemma~\ref{lem:realizability} to $\boldsymbol\pi^*$ gives a local policy $\pi''_i\in\Pi^{\mathrm{loc}}_m$ with the same distributions $d^{\pi''_i}_{i,t}=d^{\boldsymbol\pi^*}_{i,t}$ of the augmented state of agent $i$. By~\eqref{eq:G_occupancy}, it follows that $\sum_{t=0}^{H} \mathbb{E}_{\tilde s^*_{i,t} \sim d^{\boldsymbol{\pi}^*}_{i,t}}\big[ r_{i,t}^{\bar{\boldsymbol\pi}_{[i-1]}}(\tilde s^*_{i,t}) \big] = G_i^{\bar{\boldsymbol\pi}_{[i-1]}}(\pi''_i)\leq \max_{\pi_i\in\Pi^{\mathrm{loc}}_m} G_i^{\bar{\boldsymbol\pi}_{[i-1]}}(\pi_i)$. Hence,
\begin{equation}\label{eq:opt_gap}
J^* \leq J(\bar{\boldsymbol{\pi}}) + \sum_{i \in [N]} \max_{\pi_i\in\Pi^{\mathrm{loc}}_m} G_i^{\bar{\boldsymbol\pi}_{[i-1]}}(\pi_i).
\end{equation}

\emph{Step 2 (bounding the maximum local objectives by the attained ones).} Hoeffding's inequality and a union bound over the at most $(N-1)(H+1)|\mathcal S|^{m+1}$ estimated rewards give the following guarantee.

\begin{restatable}{lemma}{UniformConcentration}\label{lem:concentration}
Under~\eqref{eq:sample_complexity}, with probability at least $1-\delta$,
\begin{equation*}
 \left|\widehat r_{i,t}^{\bar{\boldsymbol\pi}_{[i-1]}}(\tilde s)
 -r_{i,t}^{\bar{\boldsymbol\pi}_{[i-1]}}(\tilde s)\right|\leq\epsilon
\end{equation*}
for all $i\geq2$, $t$, and $\tilde s\in\tilde{\mathcal S}_t$.
\end{restatable}

For the rest of the proof, we assume that the bound of Lemma~\ref{lem:concentration} holds for all $i\geq2$, $t$, and $\tilde s\in\tilde{\mathcal S}_t$, which happens with probability at least $1-\delta$. The algorithm maximizes $\widehat G_i^{\bar{\boldsymbol\pi}_{[i-1]}}$ rather than $G_i^{\bar{\boldsymbol\pi}_{[i-1]}}$. The next lemma bounds the resulting gap.

\begin{restatable}{lemma}{SequentialDpError}\label{lem:dp_error}
Let $\bar\pi_i\in\arg\max_{\pi_i\in\Pi^{\mathrm{loc}}_m}\widehat G_i^{\bar{\boldsymbol\pi}_{[i-1]}}(\pi_i)$. If
$\left|\widehat r_{i,t}^{\bar{\boldsymbol\pi}_{[i-1]}}(\tilde s)
 -r_{i,t}^{\bar{\boldsymbol\pi}_{[i-1]}}(\tilde s)\right|\leq\epsilon$
for all $t$ and $\tilde s\in\tilde{\mathcal S}_t$, then
\begin{equation*}
 G_i^{\bar{\boldsymbol\pi}_{[i-1]}}(\bar\pi_i)
 \geq\max_{\pi_i\in\Pi^{\mathrm{loc}}_m}G_i^{\bar{\boldsymbol\pi}_{[i-1]}}(\pi_i)
 -2(H+1)\epsilon.
\end{equation*}
For $i=1$, $\bar\pi_1$ attains the maximum exactly.
\end{restatable}

Applying Lemma~\ref{lem:dp_error}, each term $\max_{\pi_i\in\Pi^{\mathrm{loc}}_m} G_i^{\bar{\boldsymbol\pi}_{[i-1]}}(\pi_i)$ in \eqref{eq:opt_gap} satisfies $\max_{\pi_i\in\Pi^{\mathrm{loc}}_m} G_i^{\bar{\boldsymbol\pi}_{[i-1]}}(\pi_i) \leq G_i^{\bar{\boldsymbol\pi}_{[i-1]}}(\bar{\pi}_i) + 2(H+1)\epsilon$ for $i \geq 2$, while $\max_{\pi_i\in\Pi^{\mathrm{loc}}_m} G_i^{\bar{\boldsymbol\pi}_{[i-1]}}(\pi_i) = G_i^{\bar{\boldsymbol\pi}_{[i-1]}}(\bar{\pi}_i)$ for $i = 1$. Summing these bounds over the $N$ agents, only the $N-1$ agents with $i \geq 2$ carry the error term, so
\begin{equation}\label{eq:achieved}
  J^* \leq J(\bar{\boldsymbol{\pi}}) + \sum_{i \in [N]} G_i^{\bar{\boldsymbol\pi}_{[i-1]}}(\bar{\pi}_i) + 2(N-1)(H+1)\epsilon.
\end{equation}

\emph{Step 3 (bounding the attained local objectives by the return).} It remains to relate the sum of the attained local objectives to the achieved return $J(\bar{\boldsymbol\pi})$. The following lemma provides this connection and is the only step in which pre-window curvature enters.

\begin{restatable}{lemma}{AgentSandwich}\label{lem:agent_sandwich}
For the policy $\bar{\boldsymbol\pi}$ returned by Algorithm~\ref{alg:GWPO},
\begin{equation*}
 (1-c_m)\sum_{i \in [N]}
 G_i^{\bar{\boldsymbol\pi}_{[i-1]}}(\bar\pi_i)
 \leq J(\bar{\boldsymbol\pi})
 \leq\sum_{i \in [N]}
 G_i^{\bar{\boldsymbol\pi}_{[i-1]}}(\bar\pi_i).
\end{equation*}
\end{restatable}
If $c_m = 1$, the claimed bound reads $J(\bar{\boldsymbol{\pi}}) \geq 0$, which holds because $F$ is monotone with $F(\emptyset) = 0$. Assume $c_m<1$. The lower bound in Lemma~\ref{lem:agent_sandwich} gives
\begin{equation}\label{eq:sandwich_lower}
J(\bar{\boldsymbol{\pi}}) \geq (1-c_m) \sum_{i \in [N]} G_i^{\bar{\boldsymbol\pi}_{[i-1]}}(\bar{\pi}_i),
\end{equation}
so $\sum_{i \in [N]} G_i^{\bar{\boldsymbol\pi}_{[i-1]}}(\bar{\pi}_i) \leq J(\bar{\boldsymbol{\pi}})/(1-c_m)$. Substituting into \eqref{eq:achieved},
\begin{align*}
J^* &\leq J(\bar{\boldsymbol{\pi}}) + \frac{J(\bar{\boldsymbol{\pi}})}{1-c_m} + 2(N-1)(H+1)\epsilon\\
&= \frac{2-c_m}{1-c_m}\, J(\bar{\boldsymbol{\pi}}) + 2(N-1)(H+1)\epsilon,
\end{align*}
which rearranges to~\eqref{eq:product_guarantee}. Since the bound of Lemma~\ref{lem:concentration} holds with probability at least $1-\delta$, so does~\eqref{eq:product_guarantee}.
\end{proof}

\subsection{Proof of Lemma~\ref{lem:realizability}}
\MarginalRealizability*
\begin{proof}
Fix an agent $i$ and any $\boldsymbol\pi \in \boldsymbol\Pi_H$. Let $d^{\boldsymbol\pi}_{i,t}$ and $q^{\boldsymbol\pi}_{i,t}$ denote the distributions of agent $i$'s augmented state and of its (augmented state, action) pair at time $t$:
\begin{align*}
d^{\boldsymbol\pi}_{i,t}(\tilde s)
&= \Pr\nolimits_{\boldsymbol\pi}(\tilde s_{i,t}=\tilde s),\qquad 0\leq t\leq H,\\
q^{\boldsymbol\pi}_{i,t}(\tilde s,a)
&= \Pr\nolimits_{\boldsymbol\pi}(\tilde s_{i,t}=\tilde s,\,a_{i,t}=a),\qquad 0\leq t<H,
\end{align*}
so that $\sum_{a}q^{\boldsymbol\pi}_{i,t}(\tilde s,a) = d^{\boldsymbol\pi}_{i,t}(\tilde s)$ for $0\leq t<H$. We now construct a policy $\pi''_i\in\Pi^{\mathrm{loc}}_m$ and show by induction on $t$ that $d^{\pi''_i}_{i,t}=d^{\boldsymbol\pi}_{i,t}$. Define the local policy $\pi''_i=(\pi''_{i,0},\ldots,\pi''_{i,H-1})$. When $d^{\boldsymbol\pi}_{i,t}(\tilde s)>0$, set
\begin{equation*}
\pi''_{i,t}(a\mid\tilde s)=q^{\boldsymbol\pi}_{i,t}(\tilde s,a)\,/\,d^{\boldsymbol\pi}_{i,t}(\tilde s).
\end{equation*}
When $d^{\boldsymbol\pi}_{i,t}(\tilde s)=0$, choose $\pi''_{i,t}(\cdot\mid\tilde s)\in\mathcal D(\mathcal A)$ arbitrarily.

\emph{Base case.} At $t=0$, the history $h_{i,0}$ is empty and $s_{i,0}\sim\rho_i$. Hence, $d^{\pi''_i}_{i,0}(s,h_{i,0})=d^{\boldsymbol{\pi}}_{i,0}(s,h_{i,0})=\rho_i(s)$ for every $s\in\mathcal S$, independently of the policy.

\emph{Inductive step.} Assume $0\leq t<H$ and the inductive hypothesis $d^{\pi''_i}_{i,t} = d^{\boldsymbol{\pi}}_{i,t}$, and write $q^{\pi''_i}_{i,t}$ for agent $i$'s (augmented state, action) distribution under $\pi''_i$. If $d^{\boldsymbol\pi}_{i,t}(\tilde s)=0$, both state--action occupancies are zero by the hypothesis and the identity $\sum_a q^{\boldsymbol\pi}_{i,t}(\tilde s,a)=d^{\boldsymbol\pi}_{i,t}(\tilde s)$. Otherwise, the hypothesis and the definition of $\pi''_i$ give
\begin{equation*}
q^{\pi''_i}_{i,t}(\tilde s,a) = d^{\pi''_i}_{i,t}(\tilde s)\, \pi''_{i,t}(a\mid \tilde s) = d^{\boldsymbol{\pi}}_{i,t}(\tilde s)\, \frac{q^{\boldsymbol{\pi}}_{i,t}(\tilde s,a)}{d^{\boldsymbol{\pi}}_{i,t}(\tilde s)} = q^{\boldsymbol{\pi}}_{i,t}(\tilde s,a).
\end{equation*}
It remains to prove the equality at time $t+1$. Write $\tilde s=(s,h)$. Under decoupled dynamics, agent $i$ moves to $s'$ with probability $P_{i,t}(s'\mid s,a)$, and its augmented state becomes $(s',\varphi(h,s))$. Both depend on the joint state only through agent $i$'s own pair $(\tilde s,a)$. For an augmented state $(s',h')$ at time $t+1$, summing over the pairs that lead to it therefore gives
\begin{equation*}
d^{\boldsymbol{\pi}}_{i,t+1}(s',h') = \sum_{((s,h),a):\,\varphi(h,s)=h'} q^{\boldsymbol{\pi}}_{i,t}((s,h),a)\, P_{i,t}(s'\mid s,a).
\end{equation*}
The same relation holds for $\pi''_i$ with $q_{i,t}^{\boldsymbol\pi}$ replaced by $q_{i,t}^{\pi''_i}$. Since $q^{\pi''_i}_{i,t} = q^{\boldsymbol{\pi}}_{i,t}$, we obtain $d^{\pi''_i}_{i,t+1} = d^{\boldsymbol{\pi}}_{i,t+1}$, completing the induction.

By \eqref{eq:G_occupancy} and $d^{\pi''_i}_{i,t} = d^{\boldsymbol{\pi}}_{i,t}$, for every predecessor policy $\bar{\boldsymbol\pi}_{[i-1]}$,
\begin{equation*}
G_i^{\bar{\boldsymbol\pi}_{[i-1]}}(\pi''_i) = \sum_{t=0}^{H} \mathbb{E}_{\tilde{s}_{i,t} \sim d^{\pi''_i}_{i,t}}\big[ r_{i,t}^{\bar{\boldsymbol\pi}_{[i-1]}}(\tilde{s}_{i,t}) \big] = \sum_{t=0}^{H} \mathbb{E}_{\tilde{s}_{i,t} \sim d^{\boldsymbol{\pi}}_{i,t}}\big[ r_{i,t}^{\bar{\boldsymbol\pi}_{[i-1]}}(\tilde{s}_{i,t}) \big].
\end{equation*}
\end{proof}

\subsection{Proof of Lemma~\ref{lem:concentration}}
\UniformConcentration*
\begin{proof}
We adapt the concentration argument of \citet{chen2026}. Fix an agent $i\geq2$, a time $t$, and a history-augmented state $\tilde s_{i,t}\in\tilde{\mathcal S}_t$. The empirical reward
\begin{equation*}
\widehat{r}_{i,t}^{\bar{\boldsymbol\pi}_{[i-1]}}(\tilde{s}_{i,t}) = \frac{1}{K} \sum_{k=1}^{K} \Delta\big((t,s_{i,t}) \mid \nu_t(h_{i,t}) \cup \boldsymbol{\tau}_{[i-1]}^{k}\big)
\end{equation*}

is an average of marginal gains over $K$ newly sampled predecessor trajectories. Conditional on the previously computed policies $\bar{\boldsymbol\pi}_{[i-1]}$, these trajectories are independent and each marginal gain has mean $r_{i,t}^{\bar{\boldsymbol\pi}_{[i-1]}}(\tilde s_{i,t})$. By monotonicity and submodularity, each marginal gain lies in $[0,R]$, where $R=\max_{v\in\mathcal V}F(\{v\})$. Applying Hoeffding's inequality (Lemma~\ref{lem:hoeffding}) conditionally gives
\begin{equation*}
\Pr\!\left(
\left|\widehat r_{i,t}^{\bar{\boldsymbol\pi}_{[i-1]}}(\tilde s_{i,t})-r_{i,t}^{\bar{\boldsymbol\pi}_{[i-1]}}(\tilde s_{i,t})\right|>\epsilon
\;\middle|\;\bar{\boldsymbol\pi}_{[i-1]}
\right)
\leq 2\exp\!\left(-\frac{2K\epsilon^2}{R^2}\right).
\end{equation*}
The bound holds for every realization of the predecessor policies. Averaging over the randomness used to compute these policies gives the same bound unconditionally.
\begin{align*}
\Pr\Big( \max_{i,t,\tilde{s}_{i,t}} \big|\widehat{r}_{i,t}^{\bar{\boldsymbol\pi}_{[i-1]}}(\tilde{s}_{i,t}) - r_{i,t}^{\bar{\boldsymbol\pi}_{[i-1]}}(\tilde{s}_{i,t})\big| > \epsilon \Big)
&\leq \sum_{i,t,\tilde{s}_{i,t}} \Pr\Big( \big|\widehat{r}_{i,t}^{\bar{\boldsymbol\pi}_{[i-1]}}(\tilde{s}_{i,t}) - r_{i,t}^{\bar{\boldsymbol\pi}_{[i-1]}}(\tilde{s}_{i,t})\big| > \epsilon \Big)\\
&\leq 2(N-1)(H+1)|\mathcal{S}|^{m+1}\exp\!\Big(\!-\frac{2K\epsilon^2}{R^2}\Big).
\end{align*}
This right-hand side is at most $\delta$ if and only if $\frac{2K\epsilon^2}{R^2} \geq \ln\frac{2(N-1)(H+1)|\mathcal{S}|^{m+1}}{\delta}$, that is, if and only if $K$ satisfies \eqref{eq:sample_complexity}. Therefore, with probability at least $1-\delta$, $|\widehat{r}_{i,t}^{\bar{\boldsymbol\pi}_{[i-1]}}(\tilde{s}_{i,t}) - r_{i,t}^{\bar{\boldsymbol\pi}_{[i-1]}}(\tilde{s}_{i,t})| \leq \epsilon$ for all $i, t, \tilde{s}_{i,t}$.
\end{proof}

\subsection{Proof of Lemma~\ref{lem:dp_error}}
\SequentialDpError*
\begin{proof}
Fix the predecessor policies $\bar{\boldsymbol\pi}_{[i-1]}$. For any $\pi_i\in\Pi^{\mathrm{loc}}_m$, the exact objective~\eqref{eq:G_occupancy} is
\begin{equation*}
G_i^{\bar{\boldsymbol\pi}_{[i-1]}}(\pi_i) = \sum_{t=0}^{H} \mathbb{E}_{\tilde{s}_{i,t} \sim d^{\pi_i}_{i,t}}\big[ r_{i,t}^{\bar{\boldsymbol\pi}_{[i-1]}}(\tilde{s}_{i,t}) \big].
\end{equation*}
Algorithm~\ref{alg:GWPO} maximizes the \emph{estimated objective}
\begin{equation*}
\widehat{G}_i^{\bar{\boldsymbol\pi}_{[i-1]}}(\pi_i) = \sum_{t=0}^{H} \mathbb{E}_{\tilde{s}_{i,t} \sim d^{\pi_i}_{i,t}}\big[ \widehat{r}_{i,t}^{\bar{\boldsymbol\pi}_{[i-1]}}(\tilde{s}_{i,t}) \big],
\end{equation*}
obtained from $G_i^{\bar{\boldsymbol\pi}_{[i-1]}}$ by replacing the reward $r_{i,t}^{\bar{\boldsymbol\pi}_{[i-1]}}$ with its estimate $\widehat{r}_{i,t}^{\bar{\boldsymbol\pi}_{[i-1]}}$. The state marginal $d^{\pi_i}_{i,t}$ is identical in both objectives, since it is determined by agent $i$'s dynamics and policy alone and does not depend on the reward. We bound the difference between the true and estimated objectives:
\begin{align*}
\big| G_i^{\bar{\boldsymbol\pi}_{[i-1]}}(\pi_i) - \widehat{G}_i^{\bar{\boldsymbol\pi}_{[i-1]}}(\pi_i) \big|
&= \Big| \sum_{t=0}^{H} \mathbb{E}_{\tilde{s}_{i,t} \sim d^{\pi_i}_{i,t}}\big[ r_{i,t}^{\bar{\boldsymbol\pi}_{[i-1]}}(\tilde{s}_{i,t}) - \widehat{r}_{i,t}^{\bar{\boldsymbol\pi}_{[i-1]}}(\tilde{s}_{i,t}) \big] \Big| &\quad \text{(linearity of expectation)}\\
&\leq \sum_{t=0}^{H} \mathbb{E}_{\tilde{s}_{i,t} \sim d^{\pi_i}_{i,t}}\big[ \big| r_{i,t}^{\bar{\boldsymbol\pi}_{[i-1]}}(\tilde{s}_{i,t}) - \widehat{r}_{i,t}^{\bar{\boldsymbol\pi}_{[i-1]}}(\tilde{s}_{i,t}) \big| \big] &\quad \text{(triangle inequality)}\\
&\leq \sum_{t=0}^{H} \epsilon = (H+1)\epsilon.  &\quad \text{($|\widehat{r}_{i,t}^{\bar{\boldsymbol\pi}_{[i-1]}}(\tilde{s}_{i,t}) - r_{i,t}^{\bar{\boldsymbol\pi}_{[i-1]}}(\tilde{s}_{i,t})| \leq \epsilon$)}
\end{align*}
Since the upper bound does not depend on $\pi_i$, removing the absolute value yields two inequalities that hold for all $\pi_i \in \Pi^{\mathrm{loc}}_m$:
\begin{align}
\widehat{G}_i^{\bar{\boldsymbol\pi}_{[i-1]}}(\pi_i) - (H+1)\epsilon &\leq G_i^{\bar{\boldsymbol\pi}_{[i-1]}}(\pi_i),\label{eq:dp_a}\\
G_i^{\bar{\boldsymbol\pi}_{[i-1]}}(\pi_i) - (H+1)\epsilon &\leq \widehat{G}_i^{\bar{\boldsymbol\pi}_{[i-1]}}(\pi_i).\label{eq:dp_b}
\end{align}
Because $\bar{\pi}_i\in\arg\max_{\pi_i\in\Pi^{\mathrm{loc}}_m}\widehat{G}_i^{\bar{\boldsymbol\pi}_{[i-1]}}(\pi_i)$,
\begin{equation}
 \widehat{G}_i^{\bar{\boldsymbol\pi}_{[i-1]}}(\pi_i) \leq \widehat{G}_i^{\bar{\boldsymbol\pi}_{[i-1]}}(\bar{\pi}_i) \quad \text{for all } \pi_i\in\Pi^{\mathrm{loc}}_m.\label{eq:dp_c}
\end{equation}
Subtracting $(H+1)\epsilon$ from both sides of~\eqref{eq:dp_b} gives the left inequality below. Combining~\eqref{eq:dp_c} with~\eqref{eq:dp_a} at $\bar\pi_i$ gives the right one. Hence, for all $\pi_i\in\Pi^{\mathrm{loc}}_m$,
\begin{equation*}
  G_i^{\bar{\boldsymbol\pi}_{[i-1]}}(\pi_i) - 2(H+1)\epsilon \leq \widehat{G}_i^{\bar{\boldsymbol\pi}_{[i-1]}}(\pi_i) - (H+1)\epsilon \leq G_i^{\bar{\boldsymbol\pi}_{[i-1]}}(\bar{\pi}_i).
\end{equation*}
Taking the maximum of the left-hand side over $\pi_i\in\Pi^{\mathrm{loc}}_m$ gives the desired result:
\begin{equation*}
\max_{\pi_i\in\Pi^{\mathrm{loc}}_m} G_i^{\bar{\boldsymbol\pi}_{[i-1]}}(\pi_i) - 2(H+1)\epsilon \leq G_i^{\bar{\boldsymbol\pi}_{[i-1]}}(\bar{\pi}_i).
\end{equation*}

For $i = 1$ the reward $r_{1,t}^{\bar{\boldsymbol\pi}_{[0]}}$ is computed exactly, so $\widehat{G}_1^{\bar{\boldsymbol\pi}_{[0]}} = G_1^{\bar{\boldsymbol\pi}_{[0]}}$, the error terms vanish, and $\bar{\pi}_1$ attains the maximum exactly.
\end{proof}

\subsection{Proof of Lemma~\ref{lem:agent_sandwich}}
\AgentSandwich*
\begin{proof}
Fix a realization of the returned policy $\bar{\boldsymbol\pi}$ and a joint trajectory $\boldsymbol{\tau}$ generated under it. The trajectory determines the visited pairs $\tau_{i,t}$ and the sets $\tau_{i,t-m:t-1}$, $\tau_{i,0:t-1}$, and $\boldsymbol{\tau}_{[i-1]}$. Every relation up to the final expectation step holds for each such $\boldsymbol{\tau}$ separately, and all expectations below are over the initial states and the transition dynamics with $\bar{\boldsymbol\pi}$ held fixed. Adding the visits in lexicographic order of $(i,t)$ gives
\begin{equation}\label{eq:agent_telescoping}
F(\boldsymbol{\tau}) = \sum_{i \in [N]} \sum_{t=0}^{H} \Delta\big( \tau_{i,t} \,\big|\, \tau_{i,0:t-1} \cup \boldsymbol{\tau}_{[i-1]} \big).
\end{equation}
For each agent $i$ and time $t$, apply Lemma~\ref{lem:curvature_set_extension} with $X_t=\tau_{i,t}\subseteq\mathcal V_t$, $B=\tau_{i,0:t-m-1}$, and $Z=\tau_{i,t-m:t-1}\cup\boldsymbol{\tau}_{[i-1]}$. Since $B\cup Z=\tau_{i,0:t-1}\cup\boldsymbol{\tau}_{[i-1]}$, we obtain
\begin{equation*}
\Delta\big(\tau_{i,t} \mid \tau_{i,0:t-1} \cup \boldsymbol{\tau}_{[i-1]}\big) \geq (1-c_m)\, \Delta\big(\tau_{i,t} \mid \tau_{i,t-m:t-1} \cup \boldsymbol{\tau}_{[i-1]}\big),
\end{equation*}
while submodularity ($\tau_{i,t-m:t-1} \subseteq \tau_{i,0:t-1}$) gives
\begin{equation*}
\Delta\big(\tau_{i,t} \mid \tau_{i,0:t-1} \cup \boldsymbol{\tau}_{[i-1]}\big) \leq \Delta\big(\tau_{i,t} \mid \tau_{i,t-m:t-1} \cup \boldsymbol{\tau}_{[i-1]}\big).
\end{equation*}
For every joint trajectory $\boldsymbol{\tau}$, summing the preceding inequalities over $i\in[N]$ and $t \in \{0,\ldots,H\}$ and using~\eqref{eq:agent_telescoping} yields
\begin{equation*}
(1-c_m) \sum_{i \in [N]} \sum_{t=0}^{H} \Delta\big(\tau_{i,t} \mid \tau_{i,t-m:t-1} \cup \boldsymbol{\tau}_{[i-1]}\big) \;\leq\; F(\boldsymbol{\tau}) \;\leq\; \sum_{i \in [N]} \sum_{t=0}^{H} \Delta\big(\tau_{i,t} \mid \tau_{i,t-m:t-1} \cup \boldsymbol{\tau}_{[i-1]}\big).
\end{equation*}
Taking expectations over $\boldsymbol\tau\sim\bar{\boldsymbol\pi}$ gives $\mathbb E[F(\boldsymbol\tau)]=J(\bar{\boldsymbol\pi})$, as defined in~\eqref{eq:objective}. By linearity of expectation, each expected sum of marginal gains equals the sum of the corresponding expectations, yielding
\begin{equation} \label{eq:appendix_sandwich}
(1-c_m) \sum_{i \in [N]} \sum_{t=0}^{H} \mathbb{E}_{\boldsymbol{\tau} \sim \bar{\boldsymbol{\pi}}} \big[\Delta\big(\tau_{i,t} \mid \tau_{i,t-m:t-1} \cup \boldsymbol{\tau}_{[i-1]}\big)\big] \;\leq\; J(\bar{\boldsymbol{\pi}}) \;\leq\; \sum_{i \in [N]} \sum_{t=0}^{H} \mathbb{E}_{\boldsymbol{\tau} \sim \bar{\boldsymbol{\pi}}} \big[\Delta\big(\tau_{i,t} \mid \tau_{i,t-m:t-1} \cup \boldsymbol{\tau}_{[i-1]}\big)\big].
\end{equation}
For each agent $i$ and time $t$, the marginal gain in~\eqref{eq:appendix_sandwich} depends only on the augmented state $\tilde s_{i,t}$ and the predecessors' visited set $\boldsymbol\tau_{[i-1]}$, since $\tilde s_{i,t}$ determines $\tau_{i,t}$ and $\tau_{i,t-m:t-1}=\nu_t(h_{i,t})$. Under the fixed policy $\bar{\boldsymbol\pi}\in\boldsymbol\Pi^{\mathrm{loc}}_m$, independent initial states and decoupled dynamics imply that $\tilde s_{i,t}$ and $\boldsymbol\tau_{[i-1]}$ are independent. The expectation therefore decomposes as
\begin{align*}
\mathbb{E}_{\boldsymbol{\tau} \sim \bar{\boldsymbol\pi}}\Big[\Delta\big(\tau_{i,t} \mid \tau_{i,t-m:t-1} \cup \boldsymbol{\tau}_{[i-1]}\big)\Big]
&= \mathbb{E}_{\tilde s_{i,t} \sim d^{\bar{\pi}_i}_{i,t}}\Big[\, \mathbb{E}_{\boldsymbol{\tau}_{[i-1]} \sim \bar{\boldsymbol\pi}_{[i-1]}}\Big[ \Delta\big((t, s_{i,t}) \mid \nu_t(h_{i,t}) \cup \boldsymbol{\tau}_{[i-1]}\big) \Big] \Big]\\
&= \mathbb{E}_{\tilde s_{i,t} \sim d^{\bar{\pi}_i}_{i,t}}\big[ r_{i,t}^{\bar{\boldsymbol\pi}_{[i-1]}}(\tilde s_{i,t}) \big],
\end{align*}
where the second equality follows from the definition of the local stage reward in~\eqref{eq:local_reward}. Summing over $t \in \{0,\ldots,H\}$ and using~\eqref{eq:G_occupancy} gives
\begin{equation*}
\sum_{t=0}^{H} \mathbb E_{\boldsymbol\tau\sim\bar{\boldsymbol\pi}}
\left[
\Delta\big(\tau_{i,t} \mid \tau_{i,t-m:t-1}\cup\boldsymbol\tau_{[i-1]}\big)
\right] =
G_i^{\bar{\boldsymbol\pi}_{[i-1]}}(\bar\pi_i).
\end{equation*}
Substituting this equality for each $i\in[N]$ into~\eqref{eq:appendix_sandwich} establishes the claimed bounds.
\end{proof}

\subsection{Hoeffding's Inequality}
For completeness, we recall the concentration inequality used in the proof of Lemma~\ref{lem:concentration}.

\begin{lemma}[Hoeffding's inequality~\citep{hoeffding1963}]\label{lem:hoeffding}
Let $X_1, \dots, X_n$ be independent random variables with $X_l \in [a,b]$, and let $\bar{X} = \frac{1}{n}\sum_{l=1}^{n} X_l$ with mean $\mu = \mathbb{E}[\bar{X}]$. Then for every $\epsilon > 0$,
\begin{equation*}
\Pr\big( |\bar{X} - \mu| \geq \epsilon \big) \leq 2\exp\!\Big(\!-\frac{2n\epsilon^2}{(b-a)^2}\Big).
\end{equation*}
\end{lemma}

\section{Experimental Details}\label{app:experimental_details}

\subsection{Environment and Evaluation}\label{app:experiment_environment}

The traffic-monitoring instance uses a $20\times20$ grid and horizon $H=20$. Each UAV moves deterministically to an adjacent node or hovers. An action that leaves the grid keeps the UAV in place. Initial positions in the reported comparisons are sampled independently and uniformly over grid nodes.

The road map is generated with seed 17 and fixed across methods and algorithm seeds. Road importance is not uniform across the grid. As in real urban networks, a few congested regions require more frequent monitoring than the rest. We model this pattern with two regions of high-importance roads. For roads outside these regions, the maximum importance $w_e$ is sampled from $\mathcal N(1.5,1^2)$ and clipped to $[0,3.5]$. For roads inside these regions, $w_e$ is sampled from $\mathcal N(3.5,0.5^2)$ and clipped to $[2.5,5.5]$.

Road importance recovers according to $\alpha_\ell=1-e^{-\beta\ell}$. In the time-additive case, $\alpha_\ell=1$ for every positive gap. Returns are divided by $\sum_{e\in\mathcal E}w_e$, the sum of the maximum importances of all roads. Repeated visits after recovery can contribute additional reward, so the normalized horizon return can exceed one.

\paragraph{Policy evaluation.}
Every policy is evaluated by its return $J$ on 10,000 common initial joint states. During evaluation, \textsc{SubPO} selects the action with the highest probability for each UAV, while GWPO and GPO return deterministic policies. All reported confidence intervals are computed across algorithm seeds using Student's $t$-distribution at the 95\% confidence level, with a fixed map and evaluation states.

The memory comparison in Figure~\ref{fig:gpo_comparison}(a) uses $N=20$, $m\in\{0,1,2,3\}$, and five algorithm seeds. For each seed, we express the return improvement of GWPO over GPO as a percentage of the GPO return, and then average these percentages across seeds. In Figure~\ref{fig:budget_comparison}, each \textsc{SubPO} variant uses five training seeds, while GWPO with $m=3$ and GPO each use a single seed. In the figure, each return gap is measured relative to the GWPO return from this single seed. The confidence intervals of the gaps therefore reflect only the variation across \textsc{SubPO} seeds.

\subsection{Algorithm Settings and Hyperparameter Selection}\label{app:experiment_optimization}

\paragraph{GWPO and GPO.}
GWPO estimates local stage rewards using~\eqref{eq:empirical_reward} and optimizes each local policy by backward induction on the augmented states. GPO uses the same number of sampled trajectories as GWPO with $m=0$. GWPO computes marginal gains given full predecessor trajectories and the agent's own $m$-step history, whereas GPO uses only predecessor visits at the current time. Thus, the two methods differ even when $m=0$. Table~\ref{tab:guarantee_factors} reports the pre-window curvature and the multiplicative factor in Theorem~\ref{thm:sequential_guarantee}. These factors exclude the additive error term.

\begin{table}[b]
\centering
\caption{Pre-window curvature $c_m=e^{-\beta(m+1)}$ and the approximation factor $(1-c_m)/(2-c_m)$ of Theorem~\ref{thm:sequential_guarantee} for the traffic-monitoring reward at the listed recovery rates $\beta$ and memory lengths $m$.}
\label{tab:guarantee_factors}
\footnotesize
\setlength{\tabcolsep}{4pt}
\begin{tabular}{@{}cccc@{}}
\toprule
$\beta$ & $m$ & $c_m$ & Approximation factor \\
\midrule
0.1 & 0 & 0.905 & 0.087 \\
0.1 & 3 & 0.670 & 0.248 \\
\addlinespace
0.4 & 0 & 0.670 & 0.248 \\
0.4 & 3 & 0.202 & 0.444 \\
\addlinespace
0.8 & 0 & 0.449 & 0.355 \\
0.8 & 3 & 0.041 & 0.490 \\
\bottomrule
\end{tabular}
\end{table}

\paragraph{Number of sampled trajectories.}
We fix $\delta=10^{-6}$. For each sample count $K$, the sample-size bound~\eqref{eq:sample_complexity} then gives the tolerance
\begin{equation*}
\epsilon=R\sqrt{\frac{1}{2K}\ln\frac{2(N-1)(H+1)|\mathcal S|^{m+1}}{\delta}}.
\end{equation*}
Here $R\approx0.0105$ for the reward.

Table~\ref{tab:sample_size} reports the return $J$ for five sample counts at each of $N=6$ and $N=20$. We used GWPO with $m=3$, $\beta=0.4$, and uniform initial state distributions. Each setting was run with seeds $\{1, 2, 3\}$, and all policies were evaluated on the same $10{,}000$ initial joint states. As increasing $K$ did not noticeably change the return, we ran the remaining experiments with $\epsilon\approx3.8\times10^{-3}$ for all $N$ to reduce computation.
\begin{table}[t]
\centering
\caption{Sample counts $K$, associated $\epsilon$, and returns $J$ for GWPO with $m=3$, $\beta=0.4$, and uniform initial state distributions. We fix $\delta=10^{-6}$. Each return is the mean over seeds $\{1, 2, 3\}$ with its 95\% confidence interval. The second row gives the counts used in the remaining experiments.}
\label{tab:sample_size}
\small
\setlength{\tabcolsep}{3pt}
\begin{tabular}{@{}rccrcc@{}}
\toprule
\multicolumn{3}{c}{$N=6$} & \multicolumn{3}{c}{$N=20$}\\
\cmidrule(lr){1-3}\cmidrule(lr){4-6}
$K$ & $\epsilon$ & $J$ & $K$ & $\epsilon$ & $J$\\
\midrule
17 & $1.19\times10^{-2}$ & $0.6635\pm0.0064$ & 17 & $1.20\times10^{-2}$ & $1.8028\pm0.0116$\\
166 & $3.79\times10^{-3}$ & $0.6711\pm0.0050$ & 173 & $3.77\times10^{-3}$ & $1.8050\pm0.0058$\\
1{,}656 & $1.20\times10^{-3}$ & $0.6654\pm0.0016$ & 1{,}730 & $1.19\times10^{-3}$ & $1.8034\pm0.0067$\\
16{,}556 & $3.80\times10^{-4}$ & $0.6663\pm0.0005$ & 17{,}295 & $3.77\times10^{-4}$ & $1.8021\pm0.0055$\\
165{,}556 & $1.20\times10^{-4}$ & $0.6661\pm0.0002$ & 172{,}952 & $1.19\times10^{-4}$ & $1.8015\pm0.0009$\\
\bottomrule
\end{tabular}
\end{table}

\paragraph{\textsc{SubPO} architectures.}
Both \textsc{SubPO} variants receive one-hot encodings of the joint state and time. \textsc{SubPO-M} processes these inputs with a feedforward ReLU network, while \textsc{SubPO-NM} uses a ReLU encoder followed by a single-layer GRU with the same hidden width as the encoder. Both networks output a probability distribution over the actions of each UAV.

\paragraph{\textsc{SubPO} hyperparameter selection.}
For each team size and \textsc{SubPO} variant, we considered networks with $\{1,2\}$ ReLU layers, hidden widths $\{256,512\}$, and learning rates $\{10^{-3},3\times10^{-4}\}$. For \textsc{SubPO-NM}, these layers precede the GRU. Batch sizes were 128 for \textsc{SubPO-M} and 256 for \textsc{SubPO-NM}. An epoch is one gradient update on a fresh batch of trajectories. Table~\ref{tab:experimental_subpo_hyperparameters} reports the selected hyperparameters.

Hyperparameter selection used $\beta=0.1$ and independent nonuniform initial distributions concentrated around a fixed grid node for each UAV. Each configuration was trained for 10,000 epochs and evaluated every 500 epochs on 4,000 fixed validation initial joint states.

For each team size and variant, we selected the configuration with the highest validation return and used it for all comparisons with uniform initial positions and $\beta\in\{0.1,0.4,0.8\}$. Hyperparameter selection time is excluded from the computation-time budgets.

\begin{table}[t]
\centering
\caption{Selected neural network architectures and hyperparameters for \textsc{SubPO}. Depth denotes the number of ReLU layers; the recurrent policy additionally uses one GRU layer.}
\label{tab:experimental_subpo_hyperparameters}
\small
\begin{tabular}{@{}rrrrrrr@{}}
\toprule
& \multicolumn{3}{c}{\textsc{SubPO-M}} & \multicolumn{3}{c}{\textsc{SubPO-NM}}\\
\cmidrule(lr){2-4}\cmidrule(lr){5-7}
$N$ & Depth & Width & Learning rate & Depth & Width & Learning rate\\
\midrule
2  & 1 & 512 & $10^{-3}$ & 2 & 256 & $10^{-3}$\\
6  & 1 & 256 & $10^{-3}$ & 1 & 512 & $3\times10^{-4}$\\
10 & 1 & 512 & $10^{-3}$ & 1 & 512 & $3\times10^{-4}$\\
15 & 1 & 512 & $3\times10^{-4}$ & 1 & 512 & $10^{-3}$\\
20 & 1 & 256 & $10^{-3}$ & 1 & 512 & $10^{-3}$\\
25 & 1 & 256 & $10^{-3}$ & 1 & 512 & $10^{-3}$\\
30 & 1 & 512 & $10^{-3}$ & 1 & 512 & $3\times10^{-4}$\\
\bottomrule
\end{tabular}
\end{table}

\subsection{Computation-Time Budgets and Policy Selection}\label{app:experiment_evaluation}

\paragraph{Computation-time budgets.}
Let $T_N$ be the measured computation time of GWPO with $m=3$ for $N$ agents. GWPO and GPO run one job at a time on one AMD Ryzen 9 9950X3D CPU core, while \textsc{SubPO} uses an NVIDIA RTX PRO 6000 GPU with one CPU thread per run. For all methods, the computation time covers algorithm setup and policy optimization. For GWPO and GPO, it includes trajectory sampling and reward estimation, as well as building the augmented state spaces for GWPO. For \textsc{SubPO}, it includes training and the validations during training. Environment initialization and final test evaluation are excluded. The GWPO and GPO policies are held fixed across the budgets $T_N$, $5T_N$, and $20T_N$.

We train several \textsc{SubPO} runs in parallel, so the elapsed time of each run may overstate its computation time. We therefore convert each computation-time budget into a number of epochs, using separate calibration runs executed one at a time. For each team size and variant, the calibration process uses three seeds at $\beta=0.4$. Using the number of epochs completed by $T_N$, we estimate the counts at $5T_N$ and $20T_N$ by extrapolating the number of epochs completed between $T_N/2$ and $T_N$. We take the largest estimate across calibration seeds and add a 10\% allowance. We use the same epoch budgets for all training seeds and recovery rates, assuming that the cost of an epoch does not change across recovery rates or over training.

\paragraph{Training and policy selection.}
During each \textsc{SubPO} training run, we evaluated the greedy policy every 500 epochs on a fixed validation set of 4,000 uniformly sampled initial joint states, generated independently of the test set. We used these validation returns both to decide when to stop training and to select a policy for each epoch budget. After at least 10,000 epochs, training stopped once ten consecutive validations showed no significant improvement. An improvement was significant if the return exceeded $J_{\mathrm{ref}}$ by more than $\max\{10^{-4},10^{-3}|J_{\mathrm{ref}}|\}$. Each such improvement updated $J_{\mathrm{ref}}$ and reset the counter. 

For each computation-time budget, we selected the policy with the highest validation return reached within that budget. If a run stopped early because of the stopping criterion, we used its policy with the highest validation return for every larger budget. The selected policies were then evaluated on the test set.

\subsection{Additional Results}\label{app:experiment_results}\label{app:experiment_policies}
\begin{figure*}[t]
\centering
\setlength{\tabcolsep}{10pt}
\begin{tabular}{@{}ccc@{}}
\includegraphics{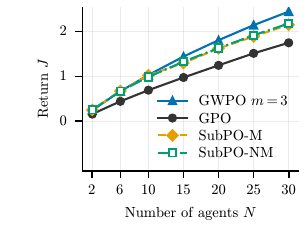} &
\includegraphics{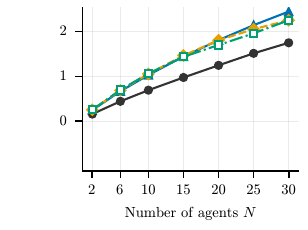} &
\includegraphics{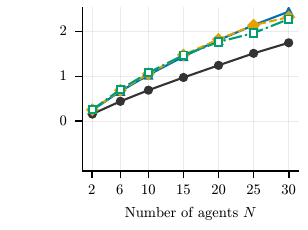} \\
{\footnotesize (a) Time budget $T_N$} &
{\footnotesize (b) Time budget $5T_N$} &
{\footnotesize (c) Time budget $20T_N$} \\
\end{tabular}
\caption{Returns of GWPO with $m=3$, GPO, and the two \textsc{SubPO} variants at $\beta=0.4$. From left to right, the allowed computation times for \textsc{SubPO} are $T_N$, $5T_N$, and $20T_N$. \textsc{SubPO} curves show means over five training seeds with 95\% confidence intervals. GWPO and GPO each use a single seed.}
\label{fig:app_budget_returns}
\end{figure*}

Figure~\ref{fig:app_budget_returns} and Table~\ref{tab:subpo_seeds} report returns $J$ at $\beta=0.4$ corresponding to the return gaps in Figure~\ref{fig:budget_comparison}.

\begin{table}[t]
\centering
\caption{Returns of the selected policies at $\beta=0.4$ with uniform initial state distributions. The \textsc{SubPO} budgets use the isolated timing calibration and 10\% allowance described above. Each \textsc{SubPO} entry is the mean over five training seeds $\pm$ the half-width of its 95\% confidence interval. GWPO with $m=3$ uses seed 1. Intervals describe training-seed variation on the fixed map and evaluation states.}
\label{tab:subpo_seeds}
\small
\setlength{\tabcolsep}{4pt}
\begin{tabular}{@{}rrlccc@{}}
\toprule
$N$ & GWPO & Policy & $T_N$ & $5T_N$ & $20T_N$ \\
\midrule
2 & 0.2484 & \textsc{SubPO-M} & $0.2526 \pm 0.0019$ & $0.2551 \pm 0.0014$ & $0.2560 \pm 0.0014$ \\
 &  & \textsc{SubPO-NM} & $0.2552 \pm 0.0008$ & $0.2570 \pm 0.0003$ & $0.2578 \pm 0.0003$ \\
\midrule
6 & 0.6691 & \textsc{SubPO-M} & $0.6747 \pm 0.0026$ & $0.6888 \pm 0.0013$ & $0.6909 \pm 0.0020$ \\
 &  & \textsc{SubPO-NM} & $0.6714 \pm 0.0027$ & $0.6979 \pm 0.0019$ & $0.7095 \pm 0.0029$ \\
\midrule
10 & 1.0312 & \textsc{SubPO-M} & $1.0254 \pm 0.0041$ & $1.0545 \pm 0.0035$ & $1.0583 \pm 0.0021$ \\
 &  & \textsc{SubPO-NM} & $0.9884 \pm 0.0054$ & $1.0657 \pm 0.0040$ & $1.0841 \pm 0.0033$ \\
\midrule
15 & 1.4427 & \textsc{SubPO-M} & $1.2984 \pm 0.0055$ & $1.4664 \pm 0.0054$ & $1.4780 \pm 0.0046$ \\
 &  & \textsc{SubPO-NM} & $1.3292 \pm 0.0028$ & $1.4374 \pm 0.0058$ & $1.4744 \pm 0.0038$ \\
\midrule
20 & 1.8076 & \textsc{SubPO-M} & $1.6249 \pm 0.0027$ & $1.8058 \pm 0.0086$ & $1.8261 \pm 0.0034$ \\
 &  & \textsc{SubPO-NM} & $1.6320 \pm 0.0013$ & $1.6998 \pm 0.0050$ & $1.7679 \pm 0.0070$ \\
\midrule
25 & 2.1429 & \textsc{SubPO-M} & $1.8882 \pm 0.0022$ & $2.0536 \pm 0.0114$ & $2.1417 \pm 0.0073$ \\
 &  & \textsc{SubPO-NM} & $1.9175 \pm 0.0024$ & $1.9608 \pm 0.0010$ & $1.9669 \pm 0.0043$ \\
\midrule
30 & 2.4447 & \textsc{SubPO-M} & $2.1526 \pm 0.0024$ & $2.2518 \pm 0.0054$ & $2.3213 \pm 0.0089$ \\
 &  & \textsc{SubPO-NM} & $2.1775 \pm 0.0040$ & $2.2517 \pm 0.0017$ & $2.2763 \pm 0.0230$ \\
\bottomrule
\end{tabular}
\end{table}
Figure~\ref{fig:app_budget_relative} extends the comparison of Figure~\ref{fig:budget_comparison} to $\beta=0.1$ and $\beta=0.8$.

\begin{figure*}[t]
\centering
\setlength{\tabcolsep}{10pt}
\begin{tabular}{@{}ccc@{}}
\multicolumn{3}{c}{\footnotesize $\beta=0.1$}\\[2pt]
\includegraphics{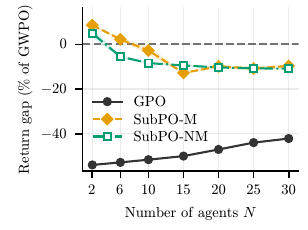} &
\includegraphics{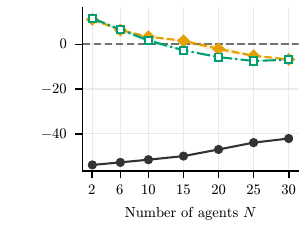} &
\includegraphics{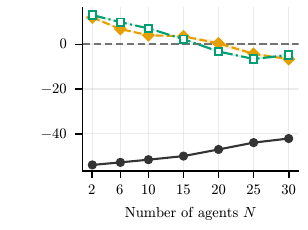} \\
{\footnotesize (a) Computation time $T_N$} &
{\footnotesize (b) Computation time $5T_N$} &
{\footnotesize (c) Computation time $20T_N$} \\[8pt]
\multicolumn{3}{c}{\footnotesize $\beta=0.8$}\\[2pt]
\includegraphics{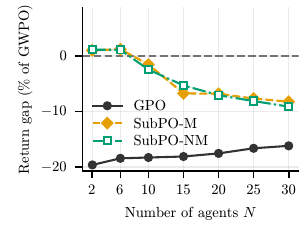} &
\includegraphics{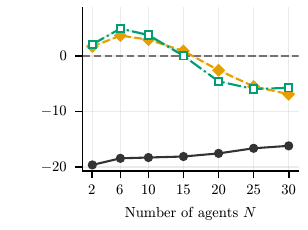} &
\includegraphics{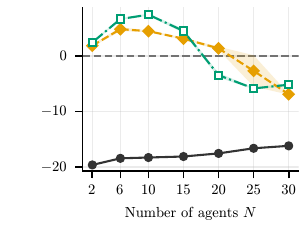} \\
{\footnotesize (d) Computation time $T_N$} &
{\footnotesize (e) Computation time $5T_N$} &
{\footnotesize (f) Computation time $20T_N$} \\
\end{tabular}
\caption{Return differences between each baseline and GWPO with $m=3$, expressed as percentages of the GWPO return, at $\beta=0.1$ (top) and $\beta=0.8$ (bottom). Negative values indicate higher returns for GWPO. From left to right, the allowed computation times for \textsc{SubPO} are $T_N$, $5T_N$, and $20T_N$. \textsc{SubPO} curves show means over five training seeds with 95\% confidence intervals. GWPO and GPO each use a single seed.}
\label{fig:app_budget_relative}
\end{figure*}

\begin{figure*}[t]
\centering
\setlength{\tabcolsep}{10pt}
\begin{tabular}{@{}ccc@{}}
\includegraphics{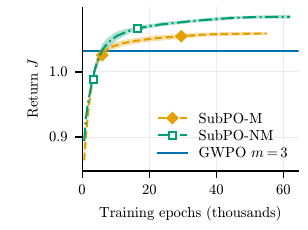} &
\includegraphics{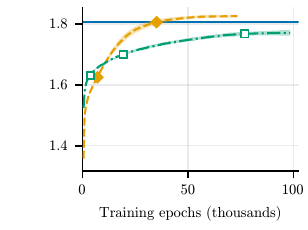} &
\includegraphics{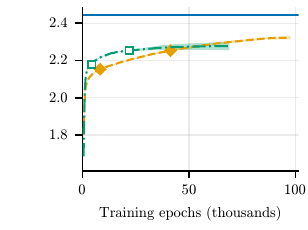} \\
{\footnotesize (a) $N=10$} &
{\footnotesize (b) $N=20$} &
{\footnotesize (c) $N=30$} \\
\end{tabular}
\caption{Training of \textsc{SubPO-M} and \textsc{SubPO-NM} at $\beta=0.4$. Curves show the highest validation return reached by each epoch, averaged over five training runs with 95\% confidence intervals. A curve ends when all five runs have stopped. Markers indicate the epoch budgets corresponding to $T_N$, $5T_N$, and $20T_N$ that fall within each curve. The horizontal line shows the test return of GWPO with $m=3$.}
\label{fig:app_subpo_training}
\end{figure*}

Figure~\ref{fig:app_subpo_training} shows the training curves. Among the runs shown, all except \textsc{SubPO-NM} at $N=20$ stopped before the $20T_N$ budget.

